%% file: _main.tex
\input{_constants}
\arxiv 

\pdfoutput=1
\documentclass[10pt,twocolumn,letterpaper]{article}
\input{cvpr_header}

\unless\ifarxiv \myexternaldocument{_supplementary} \fi

\begin{document}
\title{\paperTitle}
\author{\authorBlock}
\maketitle 

\input{00_abstract}
\input{01_intro}
\input{02_related}

\input{03_method}

\input{10_conclusion}

{\small
\bibliographystyle{ieeenat_fullname}
\bibliography{11_references}
}

\ifarxiv \clearpage \appendix \input{12_appendix} \fi

\end{document}

%% file: _constants.tex
\def\paperID{0000}
\def\confName{CVPR}
\def\confYear{2025}

\def\paperTitle{Towards Degradation-Robust Reconstruction in Generalizable NeRF}

\def\authorBlock{
    Chan Ho Park\textsuperscript{1},  \qquad
    Ka Leong Cheng\textsuperscript{1}, \qquad
    Zhicheng Wang\textsuperscript{2}, \qquad
    Qifeng Chen\textsuperscript{1} \\
    \hspace{1.0pt} \\
    \textsuperscript{1}HKUST \text{   } \textsuperscript{2}UCSD \\
}

\newif\ifreview 
\newif\ifarxiv \newcommand{\arxiv}{\arxivtrue}
\newif\ifcamera 
\newif\ifrebuttal 

%% file: cvpr_header.tex
\ifreview \usepackage[review]{cvpr} \fi
\ifarxiv \usepackage[pagenumbers]{cvpr} \fi
\ifrebuttal \usepackage[rebuttal]{cvpr} \fi
\ifcamera \usepackage{cvpr} \fi

\input{_macros}  

\usepackage{xr-hyper}

\makeatletter
\newcommand*{\addFileDependency}[1]{
  \typeout{(#1)}
  \@addtofilelist{#1}
  \IfFileExists{#1}{}{\typeout{No file #1.}}
}

\makeatother
\newcommand*{\myexternaldocument}[1]{
    \externaldocument{#1}
    \addFileDependency{#1.tex}
    \addFileDependency{#1.aux}
}

\definecolor{cvprblue}{rgb}{0.21,0.49,0.74}
\usepackage[pagebackref,breaklinks,colorlinks,allcolors=cvprblue]{hyperref}
\usepackage[capitalize]{cleveref}
\crefname{section}{Sec.}{Secs.}
\crefname{table}{Table}{Tables}
\crefname{figure}{Fig.}{Figs.}

\ifarxiv \crefname{appendix}{App.}{Apps.}
\else \crefname{appendix}{Suppl.}{Suppls.} \fi

%% file: _macros.tex
\usepackage{graphicx}	
\usepackage{amsmath}	
\usepackage{amssymb}	
\usepackage{booktabs}
\usepackage{times}
\usepackage{microtype}
\usepackage{epsfig}
\usepackage{caption}
\usepackage{float}
\usepackage{placeins}
\usepackage{color, colortbl}
\usepackage{stfloats}
\usepackage{enumitem}
\usepackage{tabularx}
\usepackage{xstring}
\usepackage{multirow}
\usepackage{xspace}
\usepackage{url}
\usepackage{subcaption}
\usepackage{xcolor}
\usepackage[hang,flushmargin]{footmisc}
\usepackage{algorithm}
\usepackage{algpseudocode}

\ifcamera \usepackage[accsupp]{axessibility} \fi

% ...

\ifarxiv  \fi

\newcommand{\R}[1]{{%
    \textbf{%
        \ifstrequal{#1}{1}{\textcolor{red}{R#1}}{%
        \ifstrequal{#1}{2}{\textcolor{blue}{R#1}}{%
        \ifstrequal{#1}{3}{\textcolor{magenta}{R#1}}{%
        \ifstrequal{#1}{4}{\textcolor{teal}{R#1}}{%
                           \textcolor{cyan}{R#1}%
        }}}}%
    }%
}}

%% file: 00_abstract.tex
\begin{abstract}

Generalizable Neural Radiance Field (GNeRF) across scenes has been proven to be an effective way to avoid per-scene optimization by representing a scene with deep image features of source images. However, despite its potential for real-world applications, there has been limited research on the robustness of GNeRFs to different types of degradation present in the source images. The lack of such research is primarily attributed to the absence of a large-scale dataset fit for training a degradation-robust generalizable NeRF model. To address this gap and facilitate investigations into the degradation robustness of 3D reconstruction tasks, we construct the \textit{Objaverse Blur Dataset}, comprising 50,000 images from over 1000 settings featuring multiple levels of blur degradation. In addition, we design a simple and model-agnostic module for enhancing the degradation robustness of GNeRFs. Specifically, by extracting 3D-aware features through a lightweight depth estimator and denoiser, the proposed module shows improvement on different popular methods in GNeRFs in terms of both quantitative and visual quality over varying degradation types and levels. Our dataset and code will be made publicly available.

\end{abstract}

%% file: 01_intro.tex
\section{Introduction}

\begin{figure}[t]
    \begin{subfigure}{0.04\linewidth}
        \parbox[][0.0cm][c]{\linewidth}{\centering\raisebox{0.7in}{\rotatebox{90}{Noise}}}
    \end{subfigure}
    \begin{subfigure}{0.95\linewidth}
        \begin{subfigure}{0.243\linewidth}
            \includegraphics[width=1\linewidth]{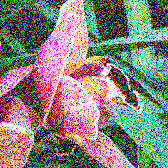}
        \end{subfigure}
        \begin{subfigure}{0.243\linewidth}
            \includegraphics[width=1\linewidth]{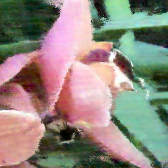}
        \end{subfigure}
        \begin{subfigure}{0.243\linewidth}
            \includegraphics[width=1\linewidth]{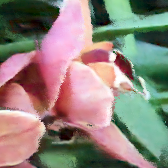}
        \end{subfigure}
        \begin{subfigure}{0.243\linewidth}
            \includegraphics[width=1\linewidth]{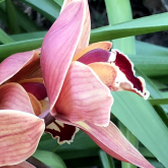}
        \end{subfigure}
    \end{subfigure}
    \begin{subfigure}{0.04\linewidth}
        \parbox[][0.0cm][c]{\linewidth}{\centering\raisebox{0.7in}{\rotatebox{90}{Blur}}}
    \end{subfigure}
    \begin{subfigure}{0.95\linewidth}
        \begin{subfigure}{0.243\linewidth}
            \includegraphics[width=1\linewidth]{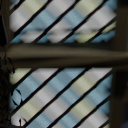}
        \end{subfigure}
        \begin{subfigure}{0.243\linewidth}
            \includegraphics[width=1\linewidth]{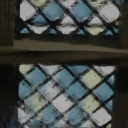}
        \end{subfigure}
        \begin{subfigure}{0.243\linewidth}
            \includegraphics[width=1\linewidth]{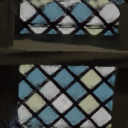}
        \end{subfigure}
        \begin{subfigure}{0.243\linewidth}
            \includegraphics[width=1\linewidth]{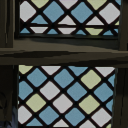}
        \end{subfigure}
    \end{subfigure}
    \begin{subfigure}{0.04\linewidth}
        \parbox[][0.0cm][c]{\linewidth}{\centering\raisebox{0.7in}{\rotatebox{90}{Blur + Noise}}}
    \end{subfigure}
    \begin{subfigure}{0.95\linewidth}
        \begin{subfigure}{0.243\linewidth}
            \includegraphics[width=1\linewidth]{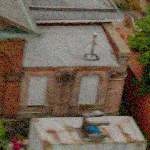}
        \end{subfigure}
        \begin{subfigure}{0.243\linewidth}
            \includegraphics[width=1\linewidth]{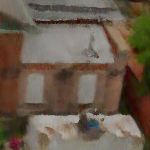}
        \end{subfigure}
        \begin{subfigure}{0.243\linewidth}
            \includegraphics[width=1\linewidth]{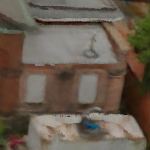}
        \end{subfigure}
        \begin{subfigure}{0.243\linewidth}
            \includegraphics[width=1\linewidth]{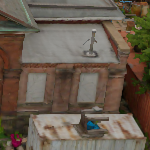}
        \end{subfigure}
    \end{subfigure}
    \begin{subfigure}{0.04\linewidth}
        \parbox[][0.0cm][c]{\linewidth}{\centering\raisebox{0.7in}{\rotatebox{90}{Adversarial}}}
    \end{subfigure}
    \begin{subfigure}{0.95\linewidth}
        \begin{subfigure}{0.243\linewidth}
            \includegraphics[width=1\linewidth]{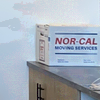}
        \end{subfigure}
        \begin{subfigure}{0.243\linewidth}
            \includegraphics[width=1\linewidth]{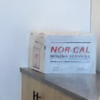}
        \end{subfigure}
        \begin{subfigure}{0.243\linewidth}
            \includegraphics[width=1\linewidth]{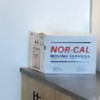}
        \end{subfigure}
        \begin{subfigure}{0.243\linewidth}
            \includegraphics[width=1\linewidth]{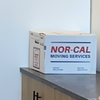}
        \end{subfigure}
    \end{subfigure}
    \begin{subfigure}{0.04\linewidth}\hfil \end{subfigure}
    \begin{subfigure}{0.95\linewidth}
        \begin{subfigure}{0.243\linewidth}\hfil Source\end{subfigure}
        \begin{subfigure}{0.243\linewidth}\hfil Baseline\end{subfigure}
        \begin{subfigure}{0.243\linewidth}\hfil Ours\end{subfigure}
        \begin{subfigure}{0.243\linewidth}\hfil Ground Truth\end{subfigure}
    \end{subfigure}
    \caption{Visual comparison of the different degradation settings against baseline. Note that there is misalignment between source and ground truth because source columns contains the source views nearby to the target view.}
    \label{fig:teaser}
\end{figure}

Robust 3D scene reconstruction under various lighting conditions is a practical task for vision and graphics applications. However, in practical image-capturing environments, image quality often degrades with noise and blur (from long exposure time, low-light environment, unstable capture, etc.), which introduces a substantial challenge to ensure the resilience of 3D rendering models to different types of degradation. Inferring the 3D geometry of a scene from a collection of 2D degraded images remains a complex and challenging task due to low image quality and the small number of captured images. 

Recently, the advent of neural radiance fields (NeRF)~\cite{mildenhall2020nerf} has significantly transformed the landscape of 3D reconstruction, making it more applicable to real-world scenarios. The limitation of vanilla NeRF has prompted extensive research focused on the optimization of both rendering~\cite{garbin2021fastnerf, hedman2021baking, reiser2021kilonerf, wizadwongsa2021nex, cao2023hexplane} and training speed \cite{muller2022instantngp, sun2021dvgo, fridovichKeil2022plenoxels}. In this process, researchers have explored various scene representations, including both explicit and implicit ones, and advanced the rendering methods in pursuit of these enhancements. From that regard, generalized NeRF (GNeRF) models~\cite{yu2021pixelnerf, chen2021mvsnerf, johari2022geonerf, suhail2022gpnr, t2023gnt, wang2021ibrnet} are a practical alternative solution to circumvent the need for scene-specific optimization. By aggregating deep image features from source images, GNeRF models construct scene representations through a single forward pass. 

Aiming to enhance the robustness of neural 3D reconstruction, several NeRF-based methods adopt source image data with haze~\cite{chen2023dehazenerf}, blur degradation~\cite{lee2023dpnerf, ma2022deblurnerf, wang2023badnerf}, low light source images~\cite{huang2023hdrnerf, jun2022hdrplenoxel, mildenhall2022hmsb}, and underwater captures~\cite{levy2023seathru, ye2022underwater}. Such NeRF-based models typically modify the rendering method based on the real physical image formation process by designing a particular renderer or module aiming to learn the latent variables for the underlying degradation. To leverage per-scene optimization methods for training in specific degradation types, prior knowledge regarding the existing degradation in the source image is necessary.


Meanwhile, there has been relatively limited exploration of GNeRF models under image degradation. Handling degradation for 3D reconstruction in a GNeRF setting~\cite{fu2023nerfool, pearl2022nan} poses a greater challenge compared to NeRF-based methods~\cite{wang2021nerf--, du2021nrf4d, mildenhall2022hmsb, wang2023benchrobust}. While methods such as~\cite{pearl2022nan, tanay2023efficientview} discuss 3D reconstruction under signal-dependent synthetic degradation with image-based rendering and multi-plane features, a more general strategy to improve robustness against degraded source images can be devised. From that regard, we propose a 3D-aware feature extraction module to enhance robustness against multiple types of degradation. The design is simple and model-agnostic, which makes it easily adaptable to various GNeRF models, broadening its potential applications in various scenarios. 

Furthermore, to address the challenge of lacking large-scale 3D reconstruction datasets for training a robust GNeRF model against blur, we construct a novel 3D reconstruction dataset with motion blur, named ~\textit{Objaverse Blur Dataset} based on Objaverse~\cite{deitke2023objaverse}. Specifically, we design an algorithm to generate the degradation using 3D models for 3D consistency, which is verified in~\cite{kar2022common3dcorr} to be crucial for ensuring compatibility with real-world applications and domain adaptation.

To this end, our contribution can be summarized as follows:
\begin{itemize}
 \item We construct the~\textit{Objaverse Blur Dataset}, the first large-scale 3D reconstruction dataset with blur degradation at multiple levels. 
 \item We propose a lightweight~\textit{3D-aware feature extractor}, with depth estimation and differentiable warping as a plugin module for GNeRF.
 \item Experimental results show that various GNeRF models with our plugin module can achieve~\textit{consistent improvement} under~\textit{multiple types of degradations}.
\end{itemize}

%% file: 02_related.tex
\section{Related Work}
\label{sec:related_works}

\noindent\textbf{Generalizable neural radiance field.} Although Neural Radiance Fields (NeRF) have proven to be a powerful implicit optimization method for modeling complex 3D scenes, the lack of generalizability in NeRF makes it impractical in real-world scenarios. Consequently, a parallel line of research in 3D reconstruction with neural rendering focuses on developing models that can generalize to new scenes without the need for fine-tuning. Generalizable NeRF (GNeRF)~\cite{yu2021pixelnerf,chen2021mvsnerf,wang2021ibrnet,johari2022geonerf,suhail2022gpnr,liu2022neuray,t2023gnt}, as inference models, map the extracted deep image features from the source images to the parameterized radiance field of the scene for volume rendering. The key variations among these methods are in their image feature aggregation, coordinate system design and rendering mechanism.

\noindent\textbf{Degradation Robustness in NeRF.} To address the presence of degradation in the source images, many degradation-robust NeRF approaches adopt a renderer that incorporate the physical image formation process. For example, approaches~\cite{martin2021nerfw,zhu2023occlusionfree,chen2022hanerf} tackle occlusion and transient objects in the source images by incorporating additional outputs capturing the radiance of the transient image. In a low-lighted source image setting, Cui~\textit{et al.}~\cite{cui2023alethnerf} incorporates the concept of the concealing field. 
Similarly, works such as~\cite{ma2022deblurnerf,lee2023dpnerf, wang2023badnerf} deal with blur degradation through the joint optimization of rays per source image and model the camera trajectory or the focal plane. With regard to GNeRF-based approaches, Pearl~\textit{et al.}~\cite{pearl2022nan} addresses source images with noise degradation. On a related note, Tanay~\textit{et al.}~\cite{tanay2023efficientview} proposes to mitigate burst noise by constructing multi-plane features (MPF) for scene representation. Another approach proposed by RaFE~\cite{zhong2024rafe} is to introduce Generative Adversarial Networks to refine the scene representation to account for the inconsistencies in source images. Lastly, NeRFool~\cite{fu2023nerfool} explores the adversarial robustness of GNeRF models by applying the strategy in~\cite{goodfellow2015explaining} and also studies the vulnerability pattern. 



\noindent\textbf{Image restoration.} Both single-image and multi-image restoration models have been proven successful under corruptions such as blur~\cite{zhong2023bit, liang2021swinir,zamir2022restormer, chen2022nafnet}, noise~\cite{midhenall2018kpn,xia2020bpn, bhat2021deeprep, dudhane2023burstormer}, rain~\cite{jiang2020derain}, mosaic~\cite{kokkinos2019iterative}, super-resolution~\cite{isobe2020videosr}, and haze~\cite{cai2016dehazenet}. Initially, CNN-based methods pioneered this domain. Subsequently, advanced architectures~\cite{liang2021swinir,zamir2022restormer, chen2022nafnet} and generative adversarial learning~\cite{kupyn2018deblurgan, kupyn2019deblurganv2} have demonstrated effectiveness in restoring realistic visual results. Some works~\cite{chen2021pretrain, liu2023degae, li2023edt, chen2023hat} devise a way to learn an adequate representation of the degraded image through a Transformer and self-supervised pre-training by low-level pretext task. For multi-frame image restoration models~\cite{midhenall2018kpn,xia2020bpn, bhat2021deeprep, dudhane2023burstormer}, one of the main challenges is to align the different frames for better restoration. This is done by predicting a blending kernel or introducing modules like Transformer for alignment. While our module shares the same purpose with the methods above, which is to represent degraded images effectively, the main difference is that the GNeRF setting makes use of camera pose information.

%% file: 03_method.tex
\begin{figure*}[t]
  \centering
  \includegraphics[width=\textwidth]{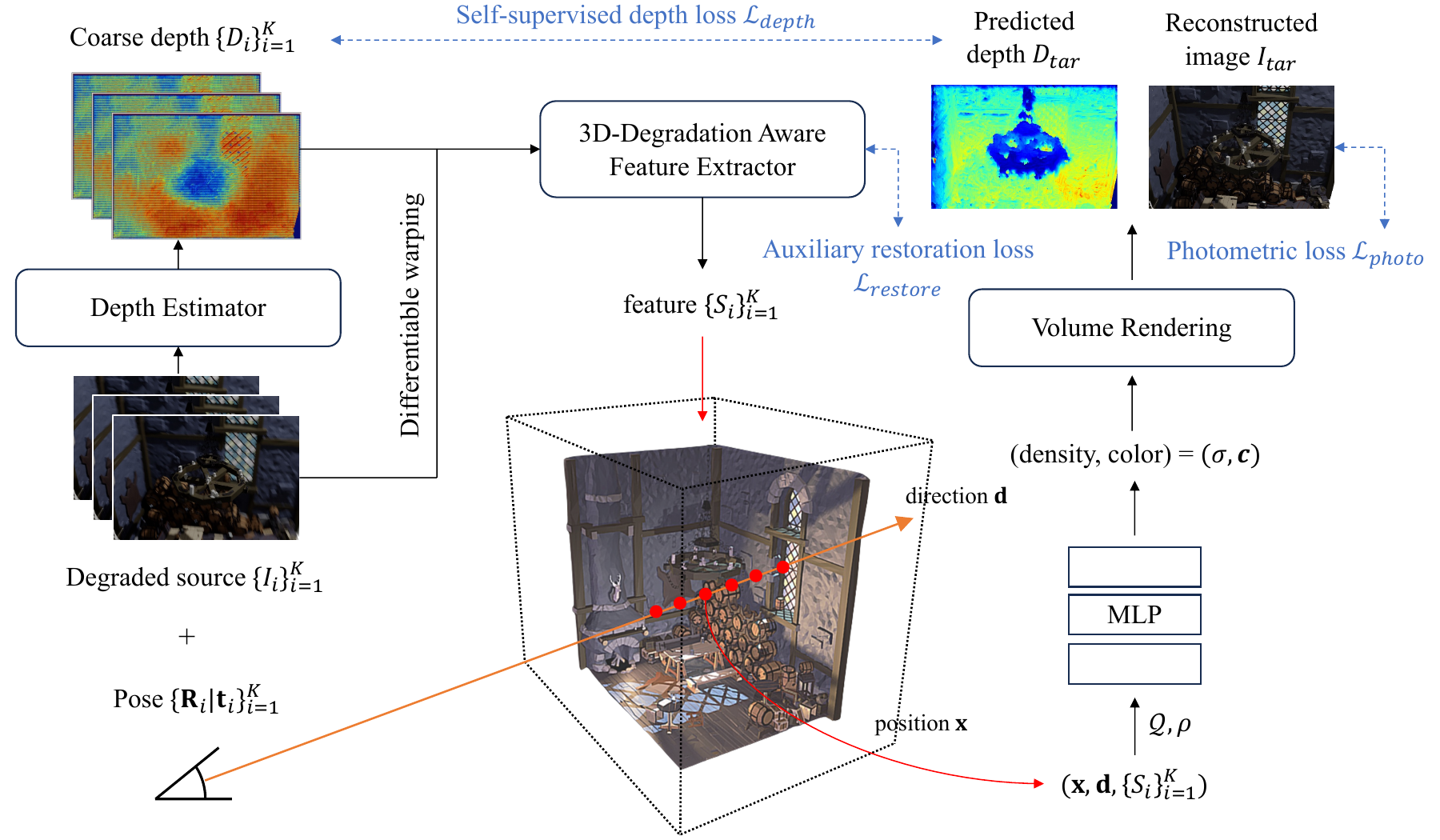}
  \caption{Training pipeline for degradation-robust GNeRF. This diagram illustrates the process for training a GNeRF that is robust to image degradation. The procedure involves a 3D-degradation-aware feature extractor $\mathcal{F}$, consisting of two steps. Initially, grouped degraded input images $\{I_{i}\}_{i=1}^{N}$ predict the corresponding depths $\{D_{i}\}_{i=1}^{N}$ through the depth estimator $\mathcal{D}$, followed by a differentiable warping. Subsequently, the 3D-degradation-aware feature extractor $\mathcal{F}$ extracts feature $\{S_{i}\}_{i=1}^{N}$ to parameterize the scene representation for the volume rendering stage.}
  \label{fig:overivew}
\end{figure*}

\section{Preliminaries}
\label{sec:gnerf}

NeRF is an implicit field that maps a query xyz coordinate $\mathbf{x} \in  \mathbb{R}^3$ and a viewing direction $\mathbf{d} \in \mathbb{R}^3$ into a radiance value $\mathbf{c} \in \mathbb{R}^3$ and density $\sigma \in \mathbb{R}$,~\textit{i.e.} $(\mathbf{c}(\mathbf{x},\mathbf{d}), \sigma(\mathbf{x})) = f(\mathbf{x}, \mathbf{d})$. Each pixel in an image is seen as a ray $\mathbf{r}$ starting from the camera origin $\mathbf{o}$ with a direction $\mathbf{d} \in \mathbb{R}^3$ which is dependent on the camera pose ${\mathbf{R}|\mathbf{t}}$ and intrinsic $\mathbf{K}$. By accumulating the radiance value from the camera origin up to a predefined far bound $t_f$, NeRF predicts a pixel value $\hat{C}(\mathbf{r})$:
\begin{equation}
\label{equ:nerf}
    \hat{C}(\mathbf{r}) = \int_0^{t_f} T(t)\sigma(\mathbf{o} + t\mathbf{d})\mathbf{c}(\mathbf{o} + t\mathbf{d}, \mathbf{d}) \enspace {\rm d}t,
\end{equation}
where $T(t) = \exp(- \int_{0}^{t} \sigma(\mathbf{o} + s\mathbf{d}) {\rm d}s)$ is the accumulated density, also referred as transmittance. NeRF is typically optimized by an $\mathcal{L}_{\rm photo}$ MSE loss between the ground truth RGB pixel $C(\mathbf{r})$ and the prediction $\hat{C}(\mathbf{r})$. Additionally, by replacing $\mathbf{c}$ in~\cref{equ:nerf} with the depth $t$, we can obtain the predicted depth. 

Under this setting, GNeRF projects the query 3D coordinate $\mathbf{x}$ to the input source images $\{I_{i}\}_{i=1}^{N}$ through the corresponding camera pose $\{\mathbf{R}_i | \mathbf{t}_i\}_{i=1}^{N}$ and intrinsics $\{\mathbf{K}_i\}_{i=1}^{N}$ for aggregating source image information. Denoting the projection from 3D space to the $i$-th image $I_i$ as $\rho_i (\cdot): \mathbb{R}^3 \rightarrow \mathbb{R}^2$ and a shared CNN feature extraction module as $\mathcal{F}$. For a query point $\mathbf{x}$, the extracted feature $\mathbf{s}_i$ from the $i$-th image $I_i$ is
\begin{equation}
    \label{equ:grid_sample}
    \mathbf{s}_i = \mathtt{GridSample}(\mathcal{F}(I_i), \rho_i(\mathbf{x})),
\end{equation}
yielding a set of image features $\mathcal{S}_\mathbf{x} = \{\mathbf{s}_i\}_{i=1}^{N}$, where $F_i$ is the extracted feature for image $I_i$ through $\mathcal{F}$.
Subsequently, GNeRF models aggregate the collected feature through an aggregation function $Q$ and decode $\mathcal{S}_\mathbf{x}$ through an implicit renderer $\mathtt{MLP}$ to obtain the radiance $\mathbf{c}$ and density value $\sigma$,~\textit{i.e.} $(\mathbf{c}, \sigma(\mathbf{x})) = \mathtt{MLP}(Q(\mathcal{S}_\mathbf{x}), \mathbf{d})$, where $\mathbf{d}$ is the viewing direction of the query point $\mathbf{x}$. Then, GNeRF models can be optimized using the same objective function as vanilla NeRF models. By training over a large number of scenes, the renderer optimizes to map the query 3D point to radiance and density value conditioned on the aggregated scene-variant feature $Q(\mathcal{S}_\mathbf{x})$, allowing generalization to new scenes.

\section{Method}

Given that there is a broad spectrum of potential degradation in the real-world application of GNeRF, source images can be degraded in both the training and inference stages. Since GNeRF models are implemented by conditioning on the projected source image features $\mathcal{S}_\mathbf{x}$, as shown in~\cite{fu2023nerfool, pearl2022nan}, GNeRF models can be vulnerable without noise-aware components. 
To this end, we introduce both the 3D and degradation awareness to the feature extractor $\mathcal{F}$.

Specifically, the proposed module comprises two main components: (i) a self-supervised depth estimator $\mathcal{D}$ ; (ii) a 3D-degradation-aware feature extractor $\mathcal{F}$ with restoration head $\mathcal{R}$. The main insight is from hypothesizing that the performance drop of GNeRF with degraded source images originates from the inaccurate geometry estimation caused by the high variance amongst the learned feature in $\mathcal{S}_\mathbf{x}$ when not specifically handling the degradation. The objective of the proposed module is two-fold. Firstly, it aims to extract visual features that are independent of degradation, ensuring accurate depth estimation. Secondly, it aims to extract features inherent in natural-looking images, facilitating effective image reconstruction. As explored in NeRFool~\cite{fu2023nerfool}, the perturbation in depth is the leading cause of a significant drop in reconstruction accuracy compared to the inaccurate texture or radiance. Therefore, the initial stage of our 3D-aware feature extractor $\mathcal{F}$ involves estimating the depth of each source image by leveraging information from nearby images. Subsequently, these nearby images are aligned and stacked together to form a composite representation, which encapsulates degradation-robust and inherent features $\mathcal{S}_\mathbf{x}$.
 
With the two-step module design, our proposed feature extractor aims not only to replace traditional convolution-based feature extractors but also to maintain comparable inference speeds to the original GNeRF model, as detailed in~\cref{sec:exp}. Additionally, it is worth noting that individual components of the 3D-degradation-aware feature extractor $\mathcal{F}$ are modular, allowing for substitution with alternative designs to accommodate variations in computational resource constraints. 

\subsection{Depth Estimation and Differentiable Warping}
\label{sec:self_supervised_depth_estimation}

Methods like~\cite{pearl2022nan} have demonstrated that denoising techniques that incorporate camera poses significantly outperform multi-frame denoising methods without the use of pose information. This is mainly through the improved alignment of the input images with larger displacement. To further leverage the camera pose information to enhance the feature extraction and denoising process, we propose a depth estimator $\mathcal{D}$ to estimate a rough depth as the first step. Specifically, we group the $K$ closest source images for each source image as in~\cref{equ:def_nearby}. Denoting the set of $K$ nearby source images for the $i$-th input image $I_i$ as $\mathcal{I}_i^K$:
\begin{equation}
\label{equ:def_nearby}
    \mathcal{I}_i^K = \{I_k : I_k \in \mathtt{NearestK}(I_i)\}.
\end{equation}
We can then compute the relative camera pose $\mathbf{R}_i^k | \mathbf{t}_i^k$ for image $I_i$ against its nearby image $I_k \in \mathcal{I}_i^K$:
\begin{equation}
    [R_i^k | t_i^k] = [R_k | t_k] \cdot [R_i | t_i]^{-1},
\end{equation}
where $\mathbf{R}_i | \mathbf{t}_i$ and $\mathbf{R}_k | \mathbf{t}_k$ denote the camera pose of $I_i$ and $I_k$, respectively. 

Subsequently, inspired by multi-view stereo (MVS)~\cite{wang2021patchmatch, yao2018mvsnet, gu2020cascademvs}, we can create the group-correlation-based matching cost volumes using the relative camera poses, which in turn estimate the dense geometric structure of the scene. Following the efficient approach of PatchMatchNet~\cite{wang2021patchmatch}, our framework incorporates a multi-scale PatchMatch algorithm for depth estimation, where we obtain the estimated depths $\{D_i\}_{i=1}^N$ for the $N$ source images. Together with the relative camera poses, our system aligns the nearby $K$ views $\mathcal{I}_i^K$ onto the source image $I_i$. Specifically, we modify the homographic warping as differentiable warping to account for the predicted depth. The mapping $\pi_i^k(\cdot) : \mathbb{R}^3 \to \mathbb{R}^3$ from a homogeneous pixel $\mathbf{p}$ in source view $I_i$ to the projected pixel in the nearby view $I_k$ is defined as follows:
\begin{equation}
\label{equ:homowarp}
    \pi_i^k(\mathbf{p})  = \mathbf{K}_k \cdot (\mathbf{R}_i^k  \cdot (\mathbf{K}_i^{-1} \cdot \mathbf{p} \cdot D_i[\mathbf{p}]) + \mathbf{t}_i^k),
\end{equation}
where $D_i[\mathbf{p}]$ denotes the depth value of pixel $\mathbf{p}$.

\begin{figure*}[t]
    \begin{subfigure}{1.0\linewidth}
        \begin{subfigure}{0.196\linewidth}
            \includegraphics[width=1\linewidth]{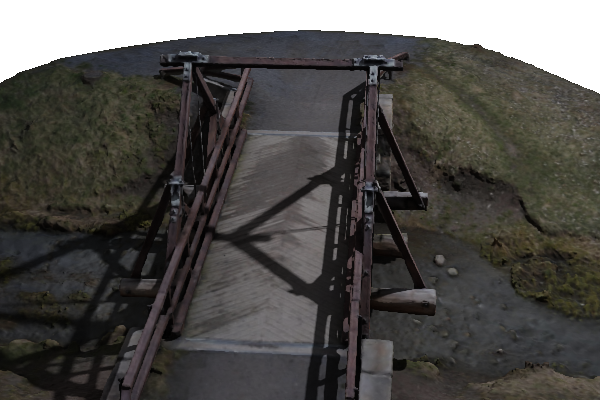}
        \end{subfigure}
        \begin{subfigure}{0.196\linewidth}
            \includegraphics[width=1\linewidth]{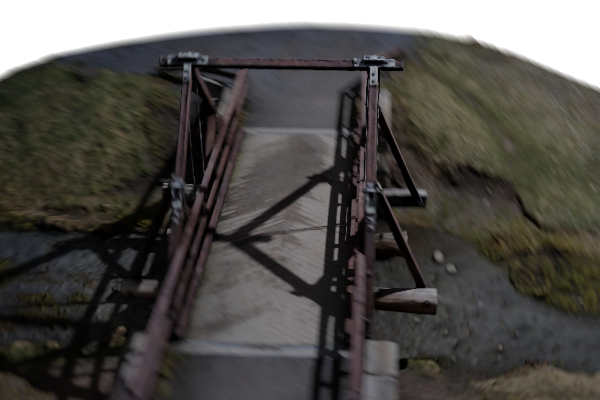}
        \end{subfigure}
        \begin{subfigure}{0.196\linewidth}
            \includegraphics[width=1\linewidth]{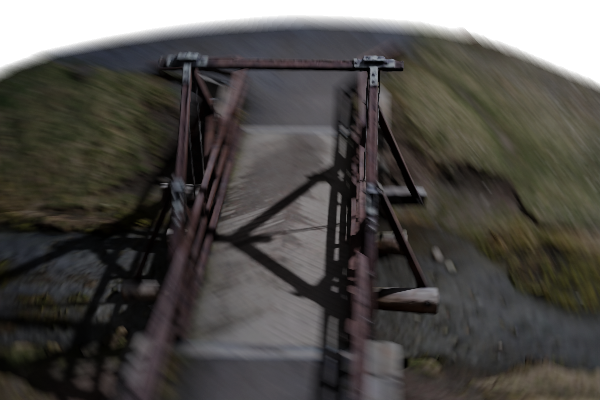}
        \end{subfigure}
        \begin{subfigure}{0.196\linewidth}
            \includegraphics[width=1\linewidth]{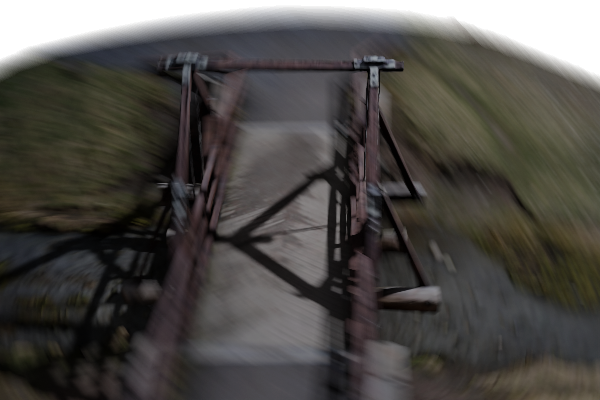}
        \end{subfigure}
        \begin{subfigure}{0.196\linewidth}
            \includegraphics[width=1\linewidth]{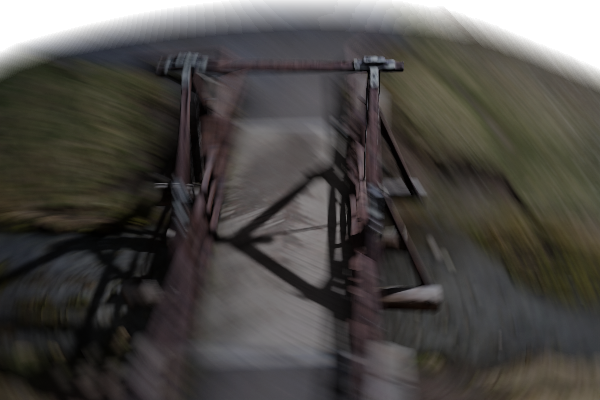}
        \end{subfigure}
    \end{subfigure}
    \begin{subfigure}{1.0\linewidth}
        \begin{subfigure}{0.196\linewidth}
            \includegraphics[width=1\linewidth]{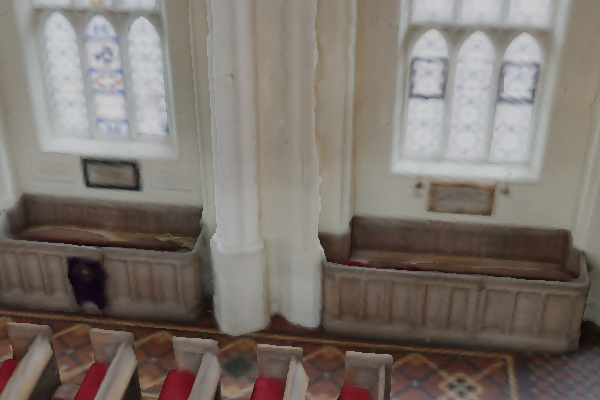}
        \end{subfigure}
        \begin{subfigure}{0.196\linewidth}
            \includegraphics[width=1\linewidth]{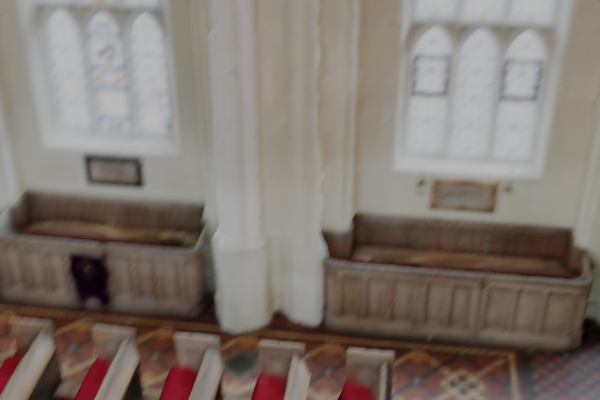}
        \end{subfigure}
        \begin{subfigure}{0.196\linewidth}
            \includegraphics[width=1\linewidth]{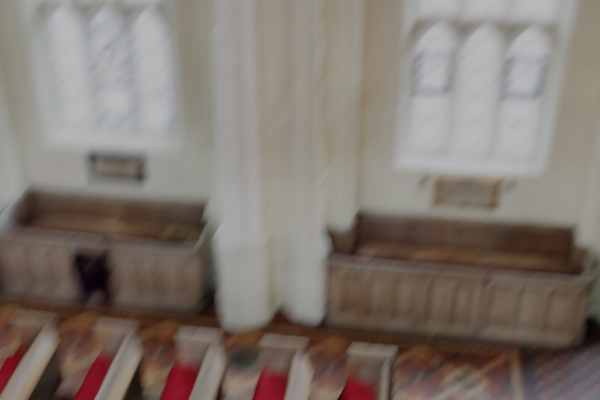}
        \end{subfigure}
        \begin{subfigure}{0.196\linewidth}
            \includegraphics[width=1\linewidth]{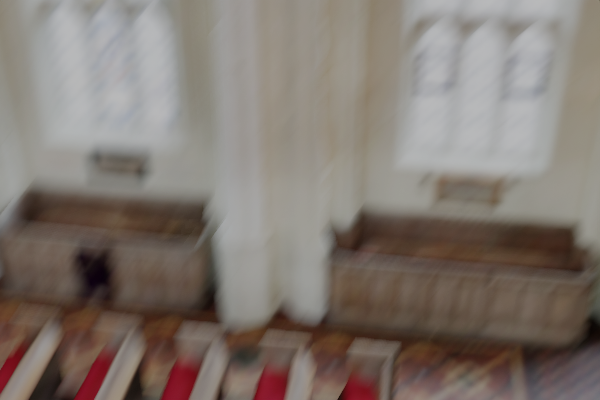}
        \end{subfigure}
        \begin{subfigure}{0.196\linewidth}
            \includegraphics[width=1\linewidth]{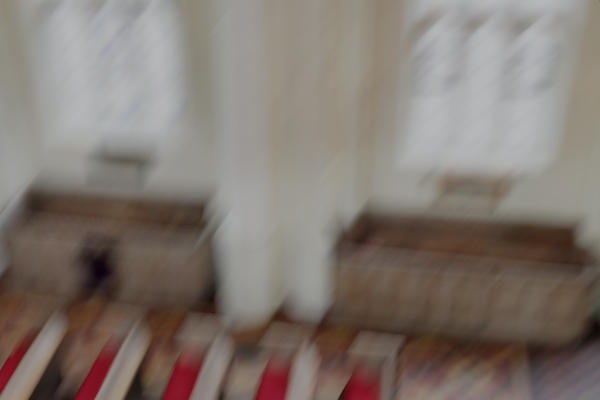}
        \end{subfigure}
    \end{subfigure}
    \begin{subfigure}{1.0\linewidth}
        \begin{subfigure}{0.196\linewidth}\hfil Without degradation\end{subfigure}
        \begin{subfigure}{0.196\linewidth}\hfil Blur level 1\end{subfigure}
        \begin{subfigure}{0.196\linewidth}\hfil Blur level 2\end{subfigure}
        \begin{subfigure}{0.196\linewidth}\hfil Blur level 3\end{subfigure}
        \begin{subfigure}{0.196\linewidth}\hfil Blur level 4\end{subfigure}
    \end{subfigure}
    \caption{Examples of Objaverse Blur Dataset at different blur levels.}
    \label{fig:objaverse-blur-dataset-examples}
\end{figure*}

\subsection{3D-Degradation-Aware Feature Extractor}

As discussed in literature such as DIP~\cite{ulyanov2017deep} and follow-up works~\cite{gandelsman2019doubledip, li2023drp, heckel2019deepdecoder}, deep image prior obtained through an auxiliary restoration module can have better capture the local-level statistics of a single natural image. Motivated by this, we introduce an auxiliary 3D-aware restoration head $\mathcal{R}$ to support the learning of our 3D-degradation-aware feature extractor $\mathcal{F}$ for better feature matching and extraction concerning multi-view consistency.

Given the source view of image $I_i$ and the nearby image $I_k$ with estimated depth map $D_k$, we can warp both $I_k$ and $D_k$ to the source view denoted as $\tilde{I}_k$ and $\tilde{D}_k$, respectively, based on ~\cref{equ:homowarp}. We concatenate all the nearby warped images $\{\tilde{I}_k\}_{k=1}^K$ and warped depths $\{\tilde{D}_k\}_{k=1}^K$ with the source image $I_i$ and its estimated depth $D_i$ to form the aligned feature $J_i$:
\begin{equation}
\label{equ:concat_nearby}
    J_i = \mathtt{Concat} (I_i, D_i, \tilde{I}_1, ..., \tilde{I}_K, \tilde{D}_1, ..., \tilde{D}_K).
\end{equation}
Inspired by multi-frame degraded-image restoration methods~\cite{xia2020bpn, bhat2021deeprep, midhenall2018kpn, dudhane2023burstormer} where source images have relatively small displacement, we leverage on the depth-aligned degraded source images $(J_i)$ through image restoration module. The intent of aligning and stacking degraded source images is to provide the image restoration module with more information from other views; the main rationale for including the predicted depth $D_i$ and the warped depths $\{\tilde{D}_i\}_{k=1}^K$ as part of the input for is that the levels of degradation within the source images such as motion blur are usually dependent on the depths~\cite{kar2022common3dcorr}. 

The aligned feature $J_i$ is then sent to the feature extractor $\mathcal{F}$ as input, which outputs a 3D-degradation-aware feature $S_i \in \mathbb{R}^{H/4 \times W/4}$. The auxiliary restoration head $\mathcal{R}$ operates in a residual manner to restore a degradation-free reconstruction $\hat{I}_i$, with an auxiliary restoration loss against the corresponding clean source image $I_i^{\rm GT}$. The features $S_i$ are used in the subsequent stage of constructing the scene representation through the aggregation function $Q$ as in~\cref{equ:grid_sample}. As the image feature shape is identical to that of the original CNN-based image feature, the proceeding steps of the volume rendering process and training strategy remain unchanged. Through such a model-agnostic feature extractor design, the proposed module can be easily plugged into any existing GNeRF under different degradation settings.

\subsection{Optimization and Loss}

During the training stage, the depth estimator $\mathcal{D}$ predicts the depth of the target view $[\mathbf{R}_{\rm tar}|\mathbf{t}_{\rm tar}]$. This prediction is based on the degraded target image $I_{\rm tar}$ and its nearby $K$ source images $\mathcal{I}_{\rm tar}^K$. As we assume the absence of ground-truth depth, the depth estimator is supervised by the target view fine depth prediction $\hat{D}_{\rm tar}$ of the GNeRF as the pseudo ground truth. With this pseudo ground truth, we detach the depth estimator from the computational graph of the main 3D reconstruction. This ensures self-supervised training of the depth estimator without requiring ground-truth depth data. We employ smooth L1 function~\cite{girshick2015fastrcnn} to the predicted depth prediction $D_{\rm tar}$:
\begin{equation}
\label{equ:depth_loss}
    \mathcal{L}_{\rm depth} = \mathtt{SmoothL1} (\hat{D}_{\rm tar}, D_{\rm tar}).
\end{equation}

Similar to~\cite{johari2022geonerf}, During the training stage, the output reconstructions of the 3D-aware restoration head $\mathcal{R}$ are supervised by the clean source images as ground truth signals are available as a supervision of the 3D reconstruction task. Formally, this loss is termed as the auxiliary restoration loss:
\begin{equation}
\label{equ:latent_reconst_loss}
    \mathcal{L}_{\rm restore} = \sum_{i=1}^K  \left\|\hat{I}_i - I_i^{\rm GT} \right\|_1,
\end{equation}
where $I_i^{\rm GT}$ is the i-$th$ clean image.

Combined with the basic 3D reconstruction photometric loss $\mathcal{L}_{\rm photo}$, the total optimization loss can be formulated as:
\begin{equation}
\label{equ:total_loss}
    \mathcal{L}_{\rm total} = \mathcal{L}_{\rm photo} + \lambda_1 \mathcal{L}_{\rm depth} + \lambda_2 \mathcal{L}_{\rm restore}
\end{equation}

\section{Objaverse Blur Dataset}
In the evaluation of robustness against degradation in GNeRF models, Pearl~\textit{et al.}~\cite{pearl2022nan} introduces a variant of the real-world forward-facing dataset called LLFF-N. This dataset comprises 43 scenes and is specifically designed to test the model performance in the presence of signal-dependent synthetic degradation, which is applied to the linear image space through inverse gamma correction. 

In practical settings, image blur is another significant and common degradation that leads to the loss of sharp details in captured images, which is widely observed in various data capture scenarios. While Ma~\textit{et al.}~\cite{ma2022deblurnerf} introduced five sets of blurry images constructed using Blender~\cite{blender}, the size of the dataset is insufficient to train a NeRF model with generalizable performance.

Motivated by these considerations, we introduce the first 3D reconstruction dataset with motion blur called the~\textit{Objaverse Blur Dataset}. Leveraging the publicly available~\textit{Objaverse}~\cite{deitke2023objaverse}, a large-scale 3D object dataset with diverse models, we simulate the real-world effects of camera motion during the image capture process. To render a single image with a specific blur level $l$ for a given 3D model, the following steps are performed:

\begin{algorithm}
    \caption{Render a blurry image from 3D model}
    \label{algo:render}
    \renewcommand{\algorithmicrequire}{\textbf{Input:}}
    \renewcommand{\algorithmicensure}{\textbf{Output:}}
    \begin{algorithmic}[1]
        \Require a 3D scene $u$ from Objaverse
        \Ensure an image with blur corruption
        \State Sample a camera position $\mathbf{p} = (r, \phi, \theta) \in \mathbb{R}^3$ distant enough from the scene or object.
        \State Sample trajectory direction $\mathbf{\delta}=(\Delta r, \Delta \phi, \Delta \theta)\in \mathbb{R}^3$.
        \State Sample a trajectory weight $w_l$ at blur level $l$ for controlling the blur strength.
        \State Uniformly sample $m$ positions $\mathbf{p}_i \sim \mathcal{U} (\mathbf{p}, \mathbf{p} + w_l\mathbf{\delta})$.
        \State Render latent images $\mathbf{x}_i$ at each position $\mathbf{p}_i$.
        \State Obtain the synthesized image $\mathbf{x} = \frac{1}{m}\sum_{i=1}^{m}\mathbf{x}_i$.
    \end{algorithmic}
\end{algorithm}

We can repeat~\cref{algo:render} to render multi-frame blurry images of a given scene with 3D consistency. Building upon the steps above, we create the Objaverse Blur Dataset, which comprises an extensive collection of over 50,000 images derived from more than 1000 unique settings obtained from a diverse set of 250 3D models. 

To discuss the details of the rendering, as the 3D models contained in Objaverse dataset~\cite{deitke2023objaverse} are diverse, the main challenge is to find the rendering configuration that aligns with the reasonable rendering in most 3D models. After tuning, the final rendering configuration is set to be as follows: 
\begin{enumerate}
  \item \textbf{Camera position $\mathbf{p}$.} We uniformly sample the camera's azimuthal and polar angles in $\phi \in (\phi_0 - \phi_{1}, \phi_0 + \phi_{1})$ and $\theta \in (\theta_0 - \theta_{1}, \theta_0 + \theta_{1})$, respectively. The parameters are set to $\phi_0 = 60^{\circ}$, $\phi_{1} = 7.5^{\circ}$, and $\theta_0$ to each of $0^{\circ}$, $90^{\circ}$, $180^{\circ}$, and $270^{\circ}$ with $\theta_{1} = 7.5^{\circ}$.
  \item \textbf{Camera trajectory $\mathbf{\delta}$.} Spherical coordinate $\Delta r, \Delta \phi, \Delta \theta \in \mathbb{R}^3$, is sampled uniformly within the range of $[\frac{1}{2}\delta_0, \delta_0]$. The parameter is set to $\delta_0 = 2.5$.
  \item \textbf{Scene dependent blur weight $w_u$.} To account for the geometric attributes of a give 3D scene $u$ at the camera position $\mathbf{p}$, such as near $n_u$, far $f_u$ and 3D bounding box $d_u$, the blur weight $w_u$ of a given 3D model is dependent on three factors: 
  \begin{enumerate}
  \item{Flatness} $F_u = \frac{f_u - n_u}{\max(d_{u})}$,
  \item{Depth Range} $R_u = \frac{f_u}{n_u}$,
  \item{Orientation} $O_u = \frac{\max(d_u)}{\min(d_u)}$.
  \end{enumerate}
  Composing all the factors, the scene dependent weight \(w_u\) is then calculated as $w_{u} = (F_u \cdot R_u \cdot O_u) ^ {1/3}$. 
  \item \textbf{Blur level $l$.} The final blur weight $w_l$ at blur level $l$ is sampled from the range of $[0.9 w_u l, 1.1 w_u l]$.
\end{enumerate}    
Following the dataset convention of Deblur-NeRF~\cite{ma2022deblurnerf}, we render $n = 34$ images at each blur level for every viewpoint sampled in the 3D model, where each image is obtained by averaging $m = 34$ latent images along the simulated camera trajectory.



\section{Evaluation}
\label{sec:exp}

\begin{figure*}[t]
    \begin{subfigure}{0.03\linewidth}
        \parbox[][0.0cm][c]{\linewidth}{\centering\raisebox{1.0in}{\rotatebox{90}{Blur}}}
    \end{subfigure}
    \begin{subfigure}{0.96\linewidth}
        \begin{subfigure}{0.24\linewidth}
            \includegraphics[page=1, width=1\linewidth]{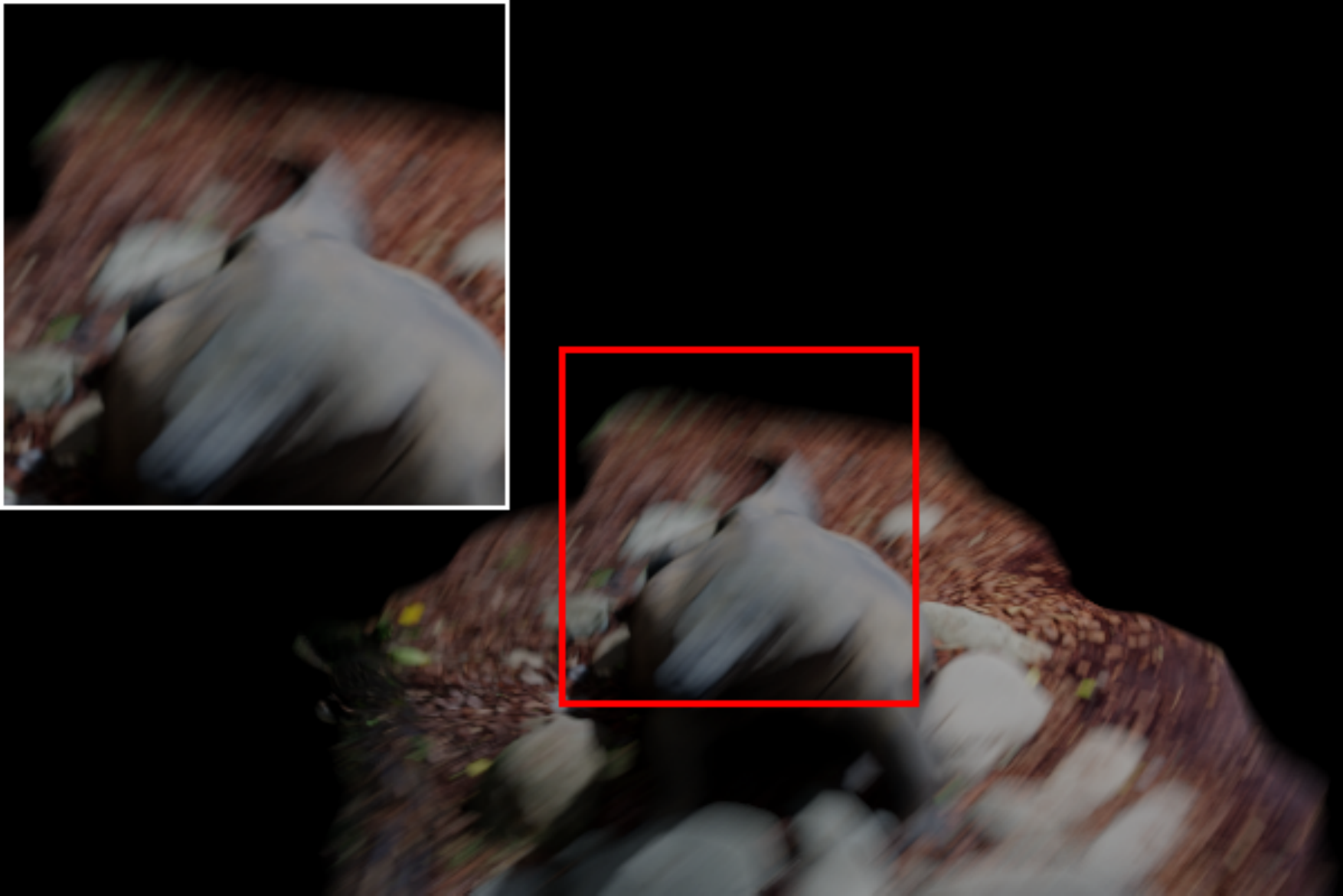}
        \end{subfigure}
        \begin{subfigure}{0.24\linewidth}
            \includegraphics[page=2, width=1\linewidth]{figures/results_row1.pdf}
        \end{subfigure}
        \begin{subfigure}{0.24\linewidth}
            \includegraphics[page=3, width=1\linewidth]{figures/results_row1.pdf}
        \end{subfigure}
        \begin{subfigure}{0.24\linewidth}
            \includegraphics[page=4, width=1\linewidth]{figures/results_row1.pdf}
        \end{subfigure}
    \end{subfigure}
    \begin{subfigure}{0.03\linewidth}\hfil \end{subfigure}
    \begin{subfigure}{0.96\linewidth}
        \begin{subfigure}{0.24\linewidth}\hfil Degraded source\end{subfigure}
        \begin{subfigure}{0.24\linewidth}\hfil GeoNeRF~\cite{t2023gnt}\end{subfigure}
        \begin{subfigure}{0.24\linewidth}\hfil Ours\end{subfigure}
        \begin{subfigure}{0.24\linewidth}\hfil Ground Truth\end{subfigure}
    \end{subfigure}
    \\[2pt]
    \begin{subfigure}{0.03\linewidth}
        \parbox[][0.0cm][c]{\linewidth}{\centering\raisebox{1.0in}{\rotatebox{90}{ Noise}}}
    \end{subfigure}
    \begin{subfigure}{0.96\linewidth}
        \begin{subfigure}{0.24\linewidth}
            \includegraphics[page=1, width=1\linewidth]{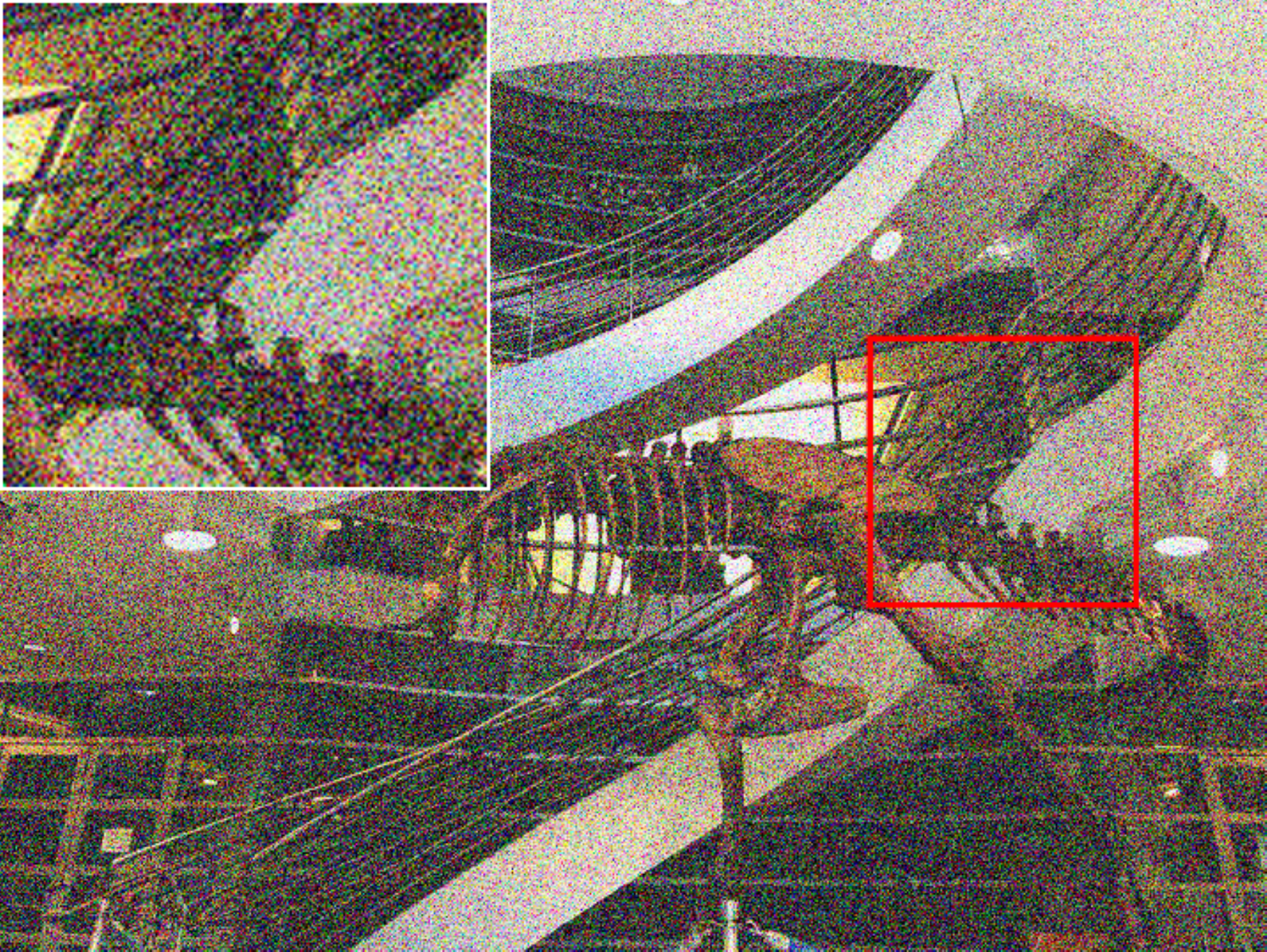}
        \end{subfigure}
        \begin{subfigure}{0.24\linewidth}
            \includegraphics[page=2, width=1\linewidth]{figures/results_row2.pdf}
        \end{subfigure}
        \begin{subfigure}{0.24\linewidth}
            \includegraphics[page=3, width=1\linewidth]{figures/results_row2.pdf}
        \end{subfigure}
        \begin{subfigure}{0.24\linewidth}
            \includegraphics[page=4, width=1\linewidth]{figures/results_row2.pdf}
        \end{subfigure}
    \end{subfigure}
    \begin{subfigure}{0.03\linewidth}
        \parbox[][0.0cm][c]{\linewidth}{\centering\raisebox{1.0in}{\rotatebox{90}{Blur + Noise}}}
    \end{subfigure}
    \begin{subfigure}{0.96\linewidth}
        \begin{subfigure}{0.24\linewidth}
            \includegraphics[page=1, width=1\linewidth]{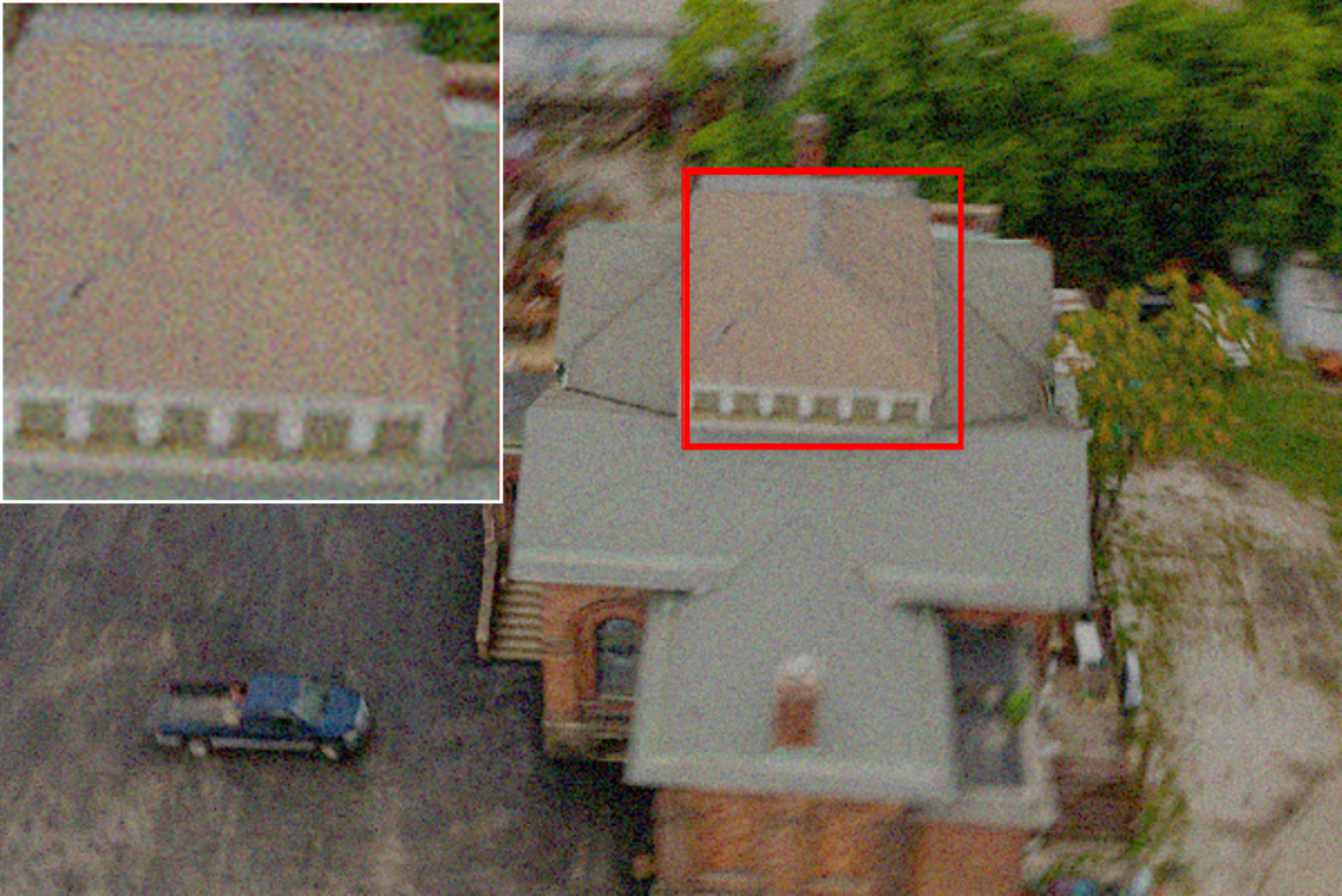}
        \end{subfigure}
        \begin{subfigure}{0.24\linewidth}
            \includegraphics[page=2, width=1\linewidth]{figures/results_row3.pdf}
        \end{subfigure}
        \begin{subfigure}{0.24\linewidth}
            \includegraphics[page=3, width=1\linewidth]{figures/results_row3.pdf}
        \end{subfigure}
        \begin{subfigure}{0.24\linewidth}
            \includegraphics[page=4, width=1\linewidth]{figures/results_row3.pdf}
        \end{subfigure}

    \end{subfigure}    
    \begin{subfigure}{0.03\linewidth}\hfil \end{subfigure}
    \begin{subfigure}{0.96\linewidth}
        \begin{subfigure}{0.24\linewidth}\hfil Degraded source\end{subfigure}
        \begin{subfigure}{0.24\linewidth}\hfil NAN~\cite{pearl2022nan}\end{subfigure}
        \begin{subfigure}{0.24\linewidth}\hfil Ours\end{subfigure}
        \begin{subfigure}{0.24\linewidth}\hfil Ground Truth\end{subfigure}
    \end{subfigure}
    \\[2pt]
    \begin{subfigure}{0.03\linewidth}
        \parbox[][0.0cm][c]{\linewidth}{\centering\raisebox{1.0in}{\rotatebox{90}{Adversarial}}}
    \end{subfigure}
    \begin{subfigure}{0.96\linewidth}
        \begin{subfigure}{0.24\linewidth}
            \includegraphics[page=1, width=1\linewidth]{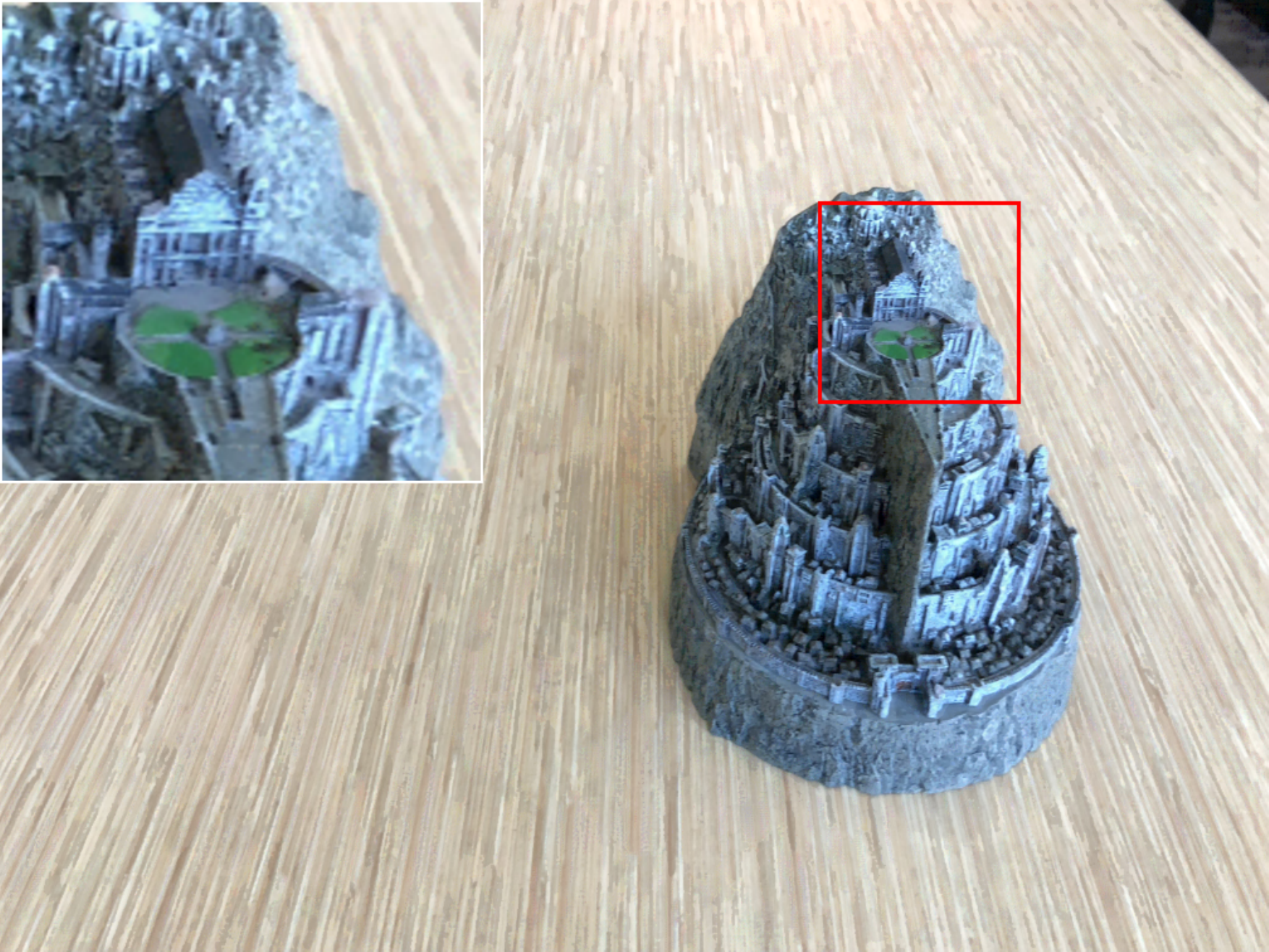}
        \end{subfigure}
        \begin{subfigure}{0.24\linewidth}
            \includegraphics[page=2, width=1\linewidth]{figures/results_row4.pdf}
        \end{subfigure}
        \begin{subfigure}{0.24\linewidth}
            \includegraphics[page=3, width=1\linewidth]{figures/results_row4.pdf}
        \end{subfigure}
        \begin{subfigure}{0.24\linewidth}
            \includegraphics[page=4, width=1\linewidth]{figures/results_row4.pdf}
        \end{subfigure}    
    \end{subfigure}
    \begin{subfigure}{0.03\linewidth}\hfil \end{subfigure}
    \begin{subfigure}{0.96\linewidth}
        \begin{subfigure}{0.24\linewidth}\hfil Degraded source\end{subfigure}
        \begin{subfigure}{0.24\linewidth}\hfil GNT~\cite{t2023gnt}\end{subfigure}
        \begin{subfigure}{0.24\linewidth}\hfil Ours\end{subfigure}
        \begin{subfigure}{0.24\linewidth}\hfil Ground Truth\end{subfigure}
    \end{subfigure}
    \caption{Qualitative comparison of GNeRFs~\cite{pearl2022nan, t2023gnt, johari2022geonerf} on different degradations and datasets~\cite{pearl2022nan, mildenhall2019llff}. Each row contains the nearest degraded source image, rendered RGB with and without our proposed module, and ground-truth target view. }
    \label{fig:comparison}
\end{figure*}

\subsection{Implementation Details}

Training of a degradation-robust GNeRF generally follows the original respective GNeRF training procedures. While employing our proposed module, two losses are added in along with the original photometric loss $\mathcal{L}_{\text{photo}}$: the auxiliary restoration loss $\mathcal{L}_{\text{restore}}$ and the self-supervised depth estimation loss $\mathcal{L}_{\text{depth}}$. These additional losses to train the 3D-degradation-aware feature extractor $\mathcal{F}$ are balanced by the corresponding weight parameters $\lambda_{\text{depth}}$ and $\lambda_{\text{rec}}$, which are empirically set to 1.0 and 0.01, respectively.  For our experiments, we have employed multi-Dconv head transposed attention (MDTA), proposed by Restormer~\cite{zamir2022restormer}, as building blocks of the 3D-aware restoration head $\mathcal{R}$.

In order to train GNeRFs with our proposed module, we have gradually annealed the auxiliary restoration loss $\mathcal{L}_{\rm restore}$ weight parameter every iteration $\lambda_{\text{restore}}^{\text{nstep}} = \text{Max} (0.01,\alpha^{\text{nstep}}) \times \lambda_{\text{restore}}$. Here, we set $\alpha = 0.99997$ and clip the minimum to 0.01 to keep the clean source image supervision to a moderate level. For training GNT~\cite{t2023gnt} for testing adversarial noise~\cite{fu2023nerfool}, we have trained and inferenced following the same setting as the original work where the number of source is 4.

For our newly constructed~\textit{Objaverse Blur Dataset}, we randomly chose 52 settings and selected test views using a similar sampling method as LLFF~\cite{mildenhall2019llff, wang2021ibrnet} to sub-sample every 16 viewpoints. As a result, our blur dataset consists of 156 test images. Furthermore, for the blur and noise degradation experimental settings, we follow the construction proposed by Pearl~\textit{et al.}~\cite{pearl2022nan} to test the 156 test images at gain levels 4,8,16,20. More details can be found in the supplementary materials.

\subsection{Results}

To demonstrate the invariance of our proposed module over the choice of GNeRF, we conduct a series of experiments with three GNeRF models: NAN~\cite{pearl2022nan}, GeoNeRF~\cite{johari2022geonerf}, and GNT~\cite{t2023gnt}. Particularly, we have selected recently proposed effective GNeRF methods of varying mechanisms. GNT~\cite{t2023gnt} is an attention-based method without an explicit rendering formula; NAN~\cite{pearl2022nan} is an image-based rendering method like IBRNet~\cite{wang2021ibrnet}; and GeoNeRF~\cite{johari2022geonerf} is a method based on an explicit cost volume scene representation like MVSNeRF~\cite{chen2021mvsnerf}.  

Degradation conditions can be classified into two categories: single degradation type, including blur degradation in~\cref{table:blur_combined_result}, noise degradation in~\cref{table:burst_result}, and adversarial degradation in~\cref{table:adversarial_result}; multiple degradations of blur and noise in~\cref{table:multi_degrade_result}. Additionally, to demonstrate the practicality of our proposed module, we include an evaluation of the real-noise image dataset in~\cref{sec:real-dataset}, stability of the depth estimation over varying degradation levels in~\cref{table:depth_stability}, and lastly impact of our proposed module on inference speed in~\cref{table:inference_speed_compare}. For conciseness, we have only included the PSNR metric and complete details of the results will be included in the supplementary materials. 

\subsection{Source Images with Single Degradation}

\begin{table}[t]
    \caption{Novel view reconstruction results of GeoNeRF~\cite{johari2022geonerf} and NAN~\cite{pearl2022nan} on the Objaverse Blur Dataset across different blur levels. Methods with * indicate the inference results of the pre-trained models trained on clean source images using pre-processed source images by Restormer~\cite{zamir2022restormer}.}
    \small
    \centering
    \resizebox{\columnwidth}{!}{%
    \begin{tabular}{ccccc}
        \toprule
        \textbf{Blur} & \textbf{Method} & \textbf{PSNR $\uparrow$} & \textbf{Method} & \textbf{PSNR $\uparrow$} \\
        \midrule
        \multirow{3}{*}{1}
            & GeoNeRF*    & 26.28                  & NAN*       & 23.73                  \\
            & GeoNeRF     & 27.96                  & NAN        & 25.59                  \\
            & + Proposed  & \textbf{28.47 (+0.51)} & + Proposed & \textbf{26.51 (+0.92)} \\
        \midrule
        \multirow{3}{*}{2}
            & GeoNeRF*    & 24.84                  & NAN*       & 22.11                  \\
            & GeoNeRF     & 26.17                  & NAN        & 24.19                  \\
            & + Proposed  & \textbf{26.73 (+0.56)} & + Proposed & \textbf{24.96 (+0.77)} \\
        \midrule
        \multirow{3}{*}{3}
            & GeoNeRF*    & 23.78                  & NAN*       & 21.14                  \\
            & GeoNeRF     & 25.12                  & NAN        & 23.28                  \\
            & + Proposed  & \textbf{25.69 (+0.57)} & + Proposed & \textbf{23.93 (+0.65)} \\
        \midrule
        \multirow{3}{*}{4}
            & GeoNeRF*    & 22.91                  & NAN*       & 20.39                  \\
            & GeoNeRF     & 24.42                  & NAN        & 22.63                  \\
            & + Proposed  & \textbf{24.98 (+0.56)} & + Proposed & \textbf{23.16 (+0.53)} \\
        \bottomrule
    \end{tabular}
    }
    \label{table:blur_combined_result}
\end{table}

\begin{table}[t]
    \caption{Novel view reconstruction results of NAN~\cite{pearl2022nan} on noise degradation (\textit{LLFF-N} dataset) across different gain levels.}
    \small
    \centering
    \begin{tabular}{lcccc}
        \toprule
        \textbf{Gain Level} & \textbf{4} & \textbf{8} & \textbf{16} & \textbf{20} \\
        \midrule
        \textbf{NAN}        & 23.85      & 23.33      & 22.36       & 21.94       \\
        \textbf{+ Proposed} & 23.90      & 23.57      & 22.89       & 22.57       \\
        \textbf{Difference} & \textbf{+0.05} & \textbf{+0.24} & \textbf{+0.53} & \textbf{+0.63} \\
        \bottomrule
    \end{tabular}
    \label{table:burst_result}
\end{table}

To assess the robustness against noise degradation, we adhered to the same setting established by NAN with 5 source images, specifically evaluating on LLFF-N dataset. Moreover, for the blur degradation setting, we have tested NAN with 7 source images and GeoNeRF with 5 source images on the Objaverse Blur Dataset while ensuring that all other components remained consistent with NAN's methodology. Additionally, for blur degradation, we compare our performance against two baselines. The first one is the clean GNeRF inferencing the pre-processed images by the state-of-the-art image restoration model Restormer~\cite{zamir2022restormer}, and the second is the GNeRF model pretrained with degraded images. For fairness, we train Restormer with the given dataset. 

Similarly, for noise degradation, we compare the performance against the original NAN model, as other baselines have been tested in the original paper. Several results are shown in~\cref{fig:comparison,table:blur_combined_result,table:burst_result} summarize the quantitative results, demonstrating consistent improvements of our model at varying blur and gain levels. Although our depth prediction seems to be smoothed compared to the original prediction, we can spot the wrong original GNeRF depth predictions, especially in terms of predicting the far surfaces.

\begin{table}[t]
    \caption{Novel view reconstruction results of GNT~\cite{t2023gnt} on different training and test methods across the~\textit{LLFF} Dataset under adversarial degradation.}
    \small
    \centering
    \resizebox{\columnwidth}{!}{%
    \begin{tabular}{llcc}
        \toprule
        \textbf{Training} & \textbf{Test} & \textbf{Proposed} & \textbf{Average PSNR $\uparrow$} \\
        \midrule
        \multirow{3}{*}{Clean} & \multirow{3}{*}{Clean} & $\times$            & 23.50 \\
                               &                        & $\checkmark$        & 24.14 \\
                               &                        & \textbf{Difference} & \textbf{+0.64} \\
        \midrule
        \multirow{3}{*}{Clean} & \multirow{3}{*}{Adversarial} & $\times$            & 17.68 \\
                               &                              & $\checkmark$        & 18.90 \\
                               &                              & \textbf{Difference} & \textbf{+1.22} \\
        \midrule
        \multirow{3}{*}{Adversarial} & \multirow{3}{*}{Clean} & $\times$            & 24.26 \\
                                     &                        & $\checkmark$        & 24.40 \\
                                     &                        & \textbf{Difference} & \textbf{+0.14} \\
        \midrule
        \multirow{3}{*}{Adversarial} & \multirow{3}{*}{Adversarial} & $\times$            & 20.05 \\
                                         &                            & $\checkmark$        & 20.22 \\
                                         &                            & \textbf{Difference} & \textbf{+0.17} \\
        \bottomrule
    \end{tabular}
    }
    \label{table:adversarial_result}
\end{table}

\begin{table}[t]
    \caption{Novel view reconstruction results of NAN~\cite{pearl2022nan} on the \textit{Objaverse Blur Dataset} across different blur levels and burst gain levels.}
    \small
    \centering
    \resizebox{\columnwidth}{!}{%
    \begin{tabular}{llccc}
        \toprule
        \textbf{Blur} & \textbf{Proposed} & \textbf{Gain 1} & \textbf{Gain 2} & \textbf{Gain 4} \\
        \midrule
        \multirow{2}{*}{1}
             & $\times$          & 23.59                & 23.47                & 23.07                \\
             & $\checkmark$      & \textbf{24.28 (+0.69)} & \textbf{24.21 (+0.74)} & \textbf{23.97 (+0.90)} \\
        \midrule
        \multirow{2}{*}{2}
             & $\times$          & 22.26                & 22.20                & 21.91                \\
             & $\checkmark$      & \textbf{22.70 (+0.44)} & \textbf{22.65 (+0.45)} & \textbf{22.39 (+0.48)} \\
        \midrule
        \multirow{2}{*}{3}
             & $\times$          & 21.48                & 21.43                & 21.20                \\
             & $\checkmark$      & \textbf{21.80 (+0.32)} & \textbf{21.73 (+0.30)} & \textbf{21.41 (+0.21)} \\
        \midrule
        \multirow{2}{*}{4}
             & $\times$          & 20.93                & 20.88                & 20.66                \\
             & $\checkmark$      & \textbf{21.17 (+0.24)} & \textbf{21.09 (+0.21)} & \textbf{20.74 (+0.08)} \\
        \bottomrule
    \end{tabular}
    }
    \label{table:multi_degrade_result}
\end{table}

To evaluate the robustness of GNeRF models with adversarial degradation in the source images, NeRFool~\cite{fu2023nerfool} compared the performance of GNT~\cite{t2023gnt} with and without adversarial training. Following the same scheme, we have trained 4 different pretrain GNT models with the two factors of whether to use adversarial training and whether to incorporate our proposed module. Based on these four models,~\cref{table:adversarial_result} shows that the proposed module is effective across all training and test combinations in enhancing the robustness of the GNeRF model for both with or without adversarial perturbation. Notably, the clean image training and clean image evaluation setups (rows 1-3) also benefit from the module. This implies that the module's integration does not merely prevent performance drop in the presence of perturbation but actively contributes to the model's ability to render scenes more accurately, even without degradation in source images. The clean image training and adversarial image evaluation settings (rows 4-6) demonstrate significant improvements. This shows that the proposed module enhances the inherent adversarial robustness of GNeRF models. Such an enhancement is crucial as it suggests that the module provides robustness against adversarial or random perturbation.


\begin{figure}[t]
    \centering
    \begin{subfigure}{1.0\linewidth}
        \begin{subfigure}{0.327\linewidth}
            \includegraphics[page=1, width=1\linewidth]{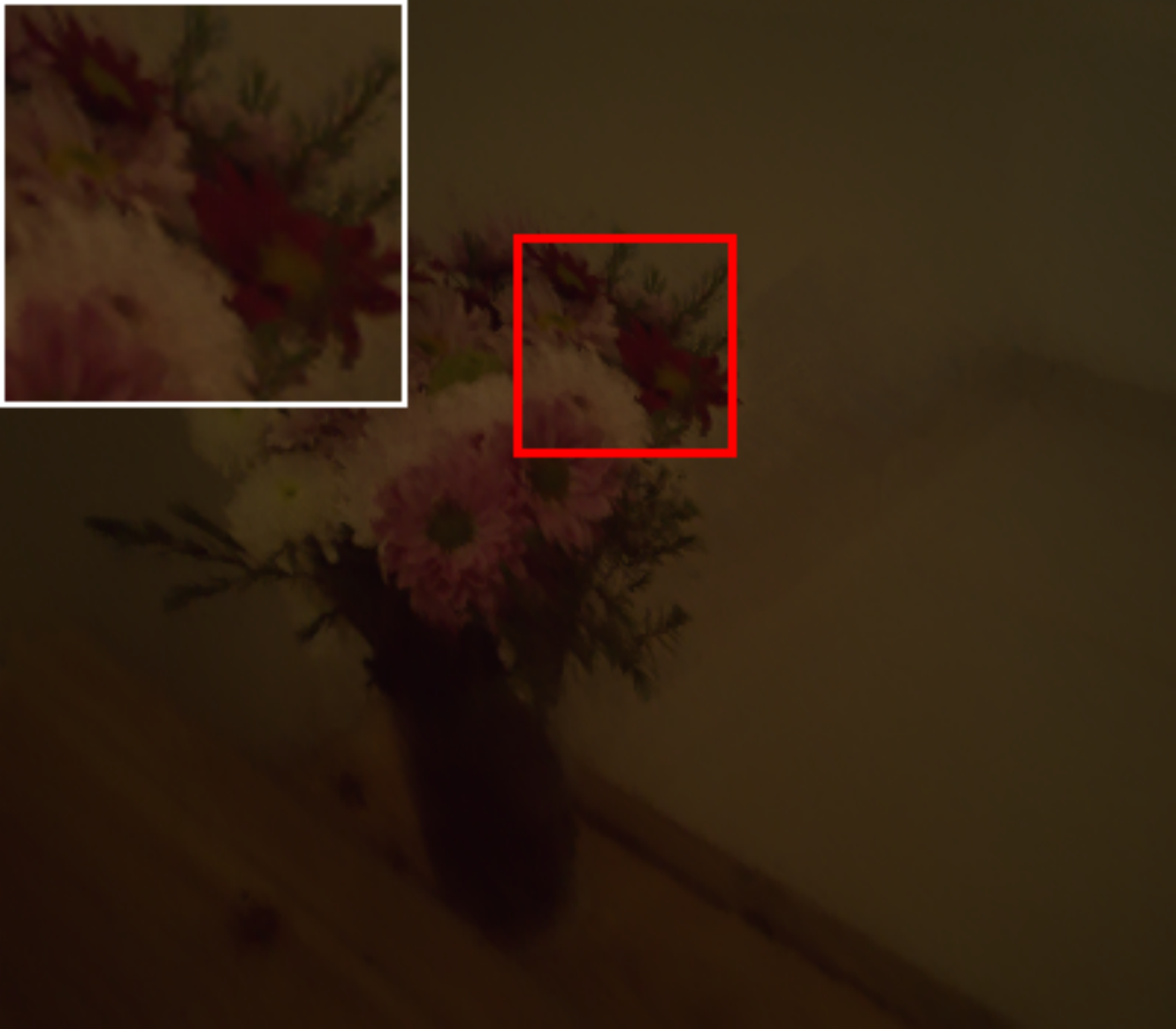}
        \end{subfigure}
        \begin{subfigure}{0.327\linewidth}
            \includegraphics[page=2, width=1\linewidth]{figures/real-result-row1.pdf}
        \end{subfigure}
        \begin{subfigure}{0.327\linewidth}
            \includegraphics[page=3, width=1\linewidth]{figures/real-result-row1.pdf}
        \end{subfigure}
    \end{subfigure}
    \begin{subfigure}{1.0\linewidth}
        \begin{subfigure}{0.327\linewidth}
            \includegraphics[page=1, width=1\linewidth]{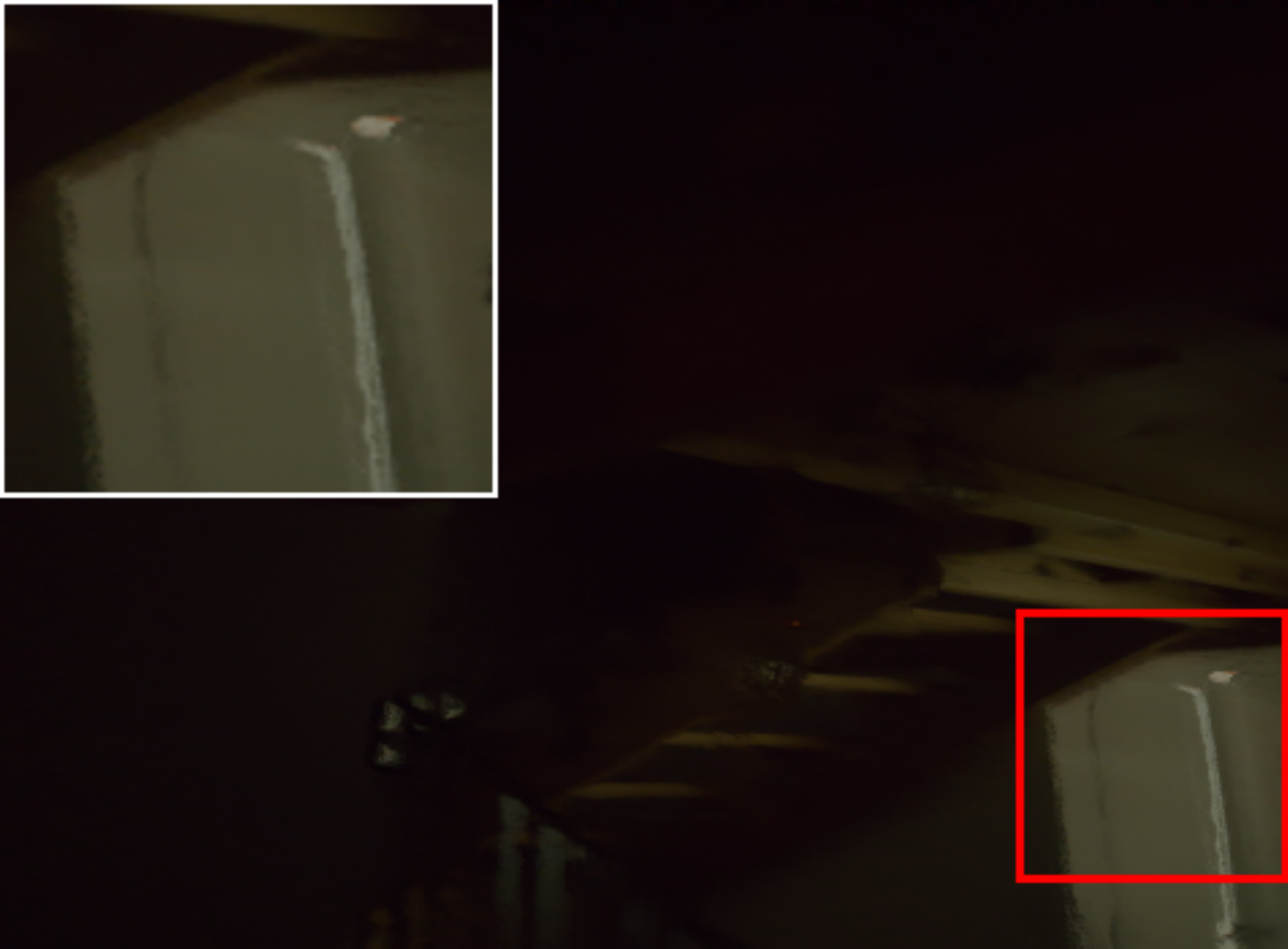}
        \end{subfigure}
        \begin{subfigure}{0.327\linewidth}
            \includegraphics[page=2, width=1\linewidth]{figures/real-result-row2.pdf}
        \end{subfigure}
        \begin{subfigure}{0.327\linewidth}
            \includegraphics[page=3, width=1\linewidth]{figures/real-result-row2.pdf}
        \end{subfigure}
    \end{subfigure}
    \begin{subfigure}{0.96\linewidth}
        \begin{subfigure}{0.32\linewidth}\hfil Baseline \end{subfigure}
        \begin{subfigure}{0.34\linewidth}\hfil Ours\end{subfigure}
        \begin{subfigure}{0.32\linewidth}\hfil Ground Truth\end{subfigure}
    \end{subfigure}
    \caption{Qualitative evaluation on real low-light noise captured with Google Pixel 4 from NAN~\cite{pearl2022nan}.}
    \label{fig:real-dataset}
\end{figure}

\subsection{Source Images with Multiple Degradations}

Table ~\ref{table:multi_degrade_result} illustrates an evaluation of the NAN’s robustness when given 5 source images with multiple degradations, specifically blur and noise. The improvement in the performance metric with the inclusion of the proposed module is consistent across all levels of blur and noise. Although the improvement weakens when the degradation strength is too large, by enhancing the model's capability across a spectrum of degradation levels, the proposed module potentially broadens the applicability and utility of GNeRF models, making them more viable for real-world, practical applications where robustness is desirable.

\subsection{Real Degraded Images}
\label{sec:real-dataset}

To assess the effectiveness of our proposed module in real applications, we qualitatively evaluate our proposed modules on captured images with real noise in the low-light environment with Google Pixel 4 provided by~\cite{pearl2022nan}. We use models trained on Objaverse Blur Dataset with noise as shown in~\cref{table:blur_combined_result}. For camera pose information, we referred to instructions described by~\cite{pearl2022nan} and used COLMAP on noisy images without any color adjustments.~\cref{fig:real-dataset} compares the rendering results with and without our proposed module.

\subsection{Analysis}

As demonstrated in NeRFool~\cite{fu2023nerfool}, stable depth prediction is crucial for robustness against degradation. In other words, given degraded source images, the model's depth prediction ideally should not change. Therefore, to evaluate the stability of depth prediction over varying blur degradation levels, we have measured the change in absolute and relative depth prediction per pixel. Here, relative depth is the absolute depth divided by the scene range. The results are summarized in~\cref{table:depth_stability}. Unsurprisingly, increments in blur degradation result in larger changes in depth per pixel, and the result also indicates that incorporating our module results in less deviation from the clean source image-based prediction. 

Additionally, since our approach replaces the conventional CNN-based feature extractor with a novel two-step process, it is essential to ensure that this modification does not adversely affect rendering speed for practicality. Therefore, to assess the impact on inference speed, we conduct tests focusing on rendering images with a resolution of $756\times1008$ pixels using an RTX 3090 graphics card. The results, presented in~\cref{table:inference_speed_compare}, detail the performance of our module in comparison to the inference speeds of GNeRF models evaluated in our experiments. Notably, the inference speeds observed with our module are commensurate with those recorded for the tested GNeRF models.

\begin{table}[t]
    \caption{Analysis of depth prediction over varying blur degradation compared to the clean source image prediction.}
    \small
    \centering
    \resizebox{\columnwidth}{!}{%
    \begin{tabular}{l l c c c c}
        \toprule
        \textbf{Blur} & \textbf{Proposed} & \textbf{$\Delta$Abs. Depth $\downarrow$} & \textbf{Diff.} & \textbf{$\Delta$Rel. Depth $\downarrow$} & \textbf{Diff.} \\
        \midrule
        \multirow{2}{*}{0} 
            & $\times$     & 0.0   & -          & 0.0   & -          \\
            & $\checkmark$ & 0.0   &            & 0.0   &            \\
        \midrule
        \multirow{2}{*}{1} 
            & $\times$     & 0.394 & -28.17\%   & 0.068 & -26.47\%   \\
            & $\checkmark$ & 0.283 &            & 0.050 &            \\
        \midrule
        \multirow{2}{*}{2} 
            & $\times$     & 0.474 & -32.91\%   & 0.080 & -32.50\%   \\
            & $\checkmark$ & 0.318 &            & 0.054 &            \\
        \midrule
        \multirow{2}{*}{3} 
            & $\times$     & 0.528 & -33.14\%   & 0.085 & -34.12\%   \\
            & $\checkmark$ & 0.353 &            & 0.056 &            \\
        \midrule
        \multirow{2}{*}{4} 
            & $\times$     & 0.562 & -32.56\%   & 0.089 & -34.83\%   \\
            & $\checkmark$ & 0.379 &            & 0.058 &            \\
        \bottomrule
    \end{tabular}
    }
    \label{table:depth_stability}
\end{table}

\begin{table}[t]
    \caption{Image rendering time (in seconds) of different GNeRFs with and without 3D-degradation-aware feature extractor $\mathcal{F}$.}
    \small
    \centering
    \resizebox{\columnwidth}{!}{%
    \begin{tabular}{l c c c c}
        \toprule
        \textbf{GNeRF} & \textbf{\# Source} & \textbf{Original} & \textbf{Proposed} & \textbf{Difference} \\
        \midrule
        \multirow{2}{*}{GeoNeRF}
            & 3 & 62.04 & 62.27 & +0.23 \\
            & 5 & 95.79 & 94.81 & -0.98 \\
        \midrule
        \multirow{2}{*}{GNT}
            & 3 & 75.54 & 76.11 & +0.57 \\
            & 5 & 90.08 & 90.38 & +0.31 \\
        \midrule
        \multirow{2}{*}{NAN}
            & 3 & 54.34 & 44.98 & -9.36 \\
            & 5 & 76.95 & 68.80 & -8.15 \\
        \bottomrule
    \end{tabular}
    }
    \label{table:inference_speed_compare}
\end{table}

%% file: 10_conclusion.tex
\section{Conclusion}

We present a model-agnostic module that can be easily incorporated into GNeRF training pipelines in order to enhance the degradation robustness. The proposed 3D-degradation-aware feature extractor has demonstrated effectiveness in enhancing the quality of the rendered images and stability of depth prediction under conditions where source images contain various or multiple types of degradations. Our plugin module consists of a two-step process: depth prediction and latent image reconstruction. Each component of the module is supervised and self-supervised from the clean source images and GNeRF depth prediction, respectively. Our proposed plugin module is modular in a way such that other designs of the two components can be easily explored. Similarly, the capacity of the modules can be adjusted based on the computational overhead of the deploy environment. Additionally, in order to evaluate the robustness against various degradations, we have constructed a novel 3D reconstruction dataset named Objaverse Blur Dataset, which simulates realistic blur image captures at different levels. 
Lastly, possible follow-up research directions include utilizing a physical image formulation model to enhance the robustness of GNeRF models and making these modules more interpretable through other scene representations.

%% file: 12_appendix.tex
\section{Appendix Section}
\label{sec:appendix_section}

\begin{figure*}[t]
    \caption{Histogram of scene rang eand dimension of Objaverse Dataset.}
    \begin{subfigure}{0.48\linewidth}
        \includegraphics[width=1\linewidth]{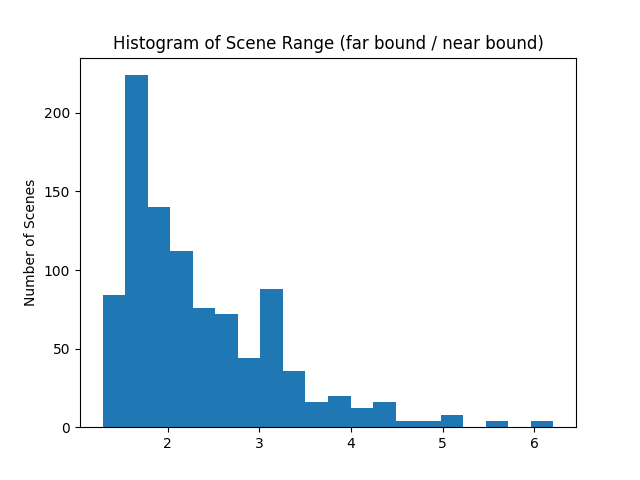}
    \end{subfigure}
    \begin{subfigure}{0.48\linewidth}
        \includegraphics[width=1\linewidth]{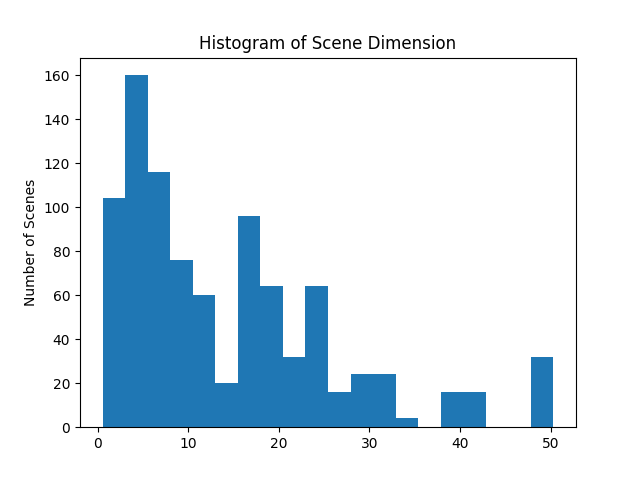}
    \end{subfigure}
        
    \label{fig:histogram_scene}
\end{figure*}

\begin{figure*}[t]
    \caption{Histogram of blur weights at different levels of \textit{Objaverse Blur} dataset.}
    \begin{subfigure}{0.24\linewidth}
        \includegraphics[width=1\linewidth]{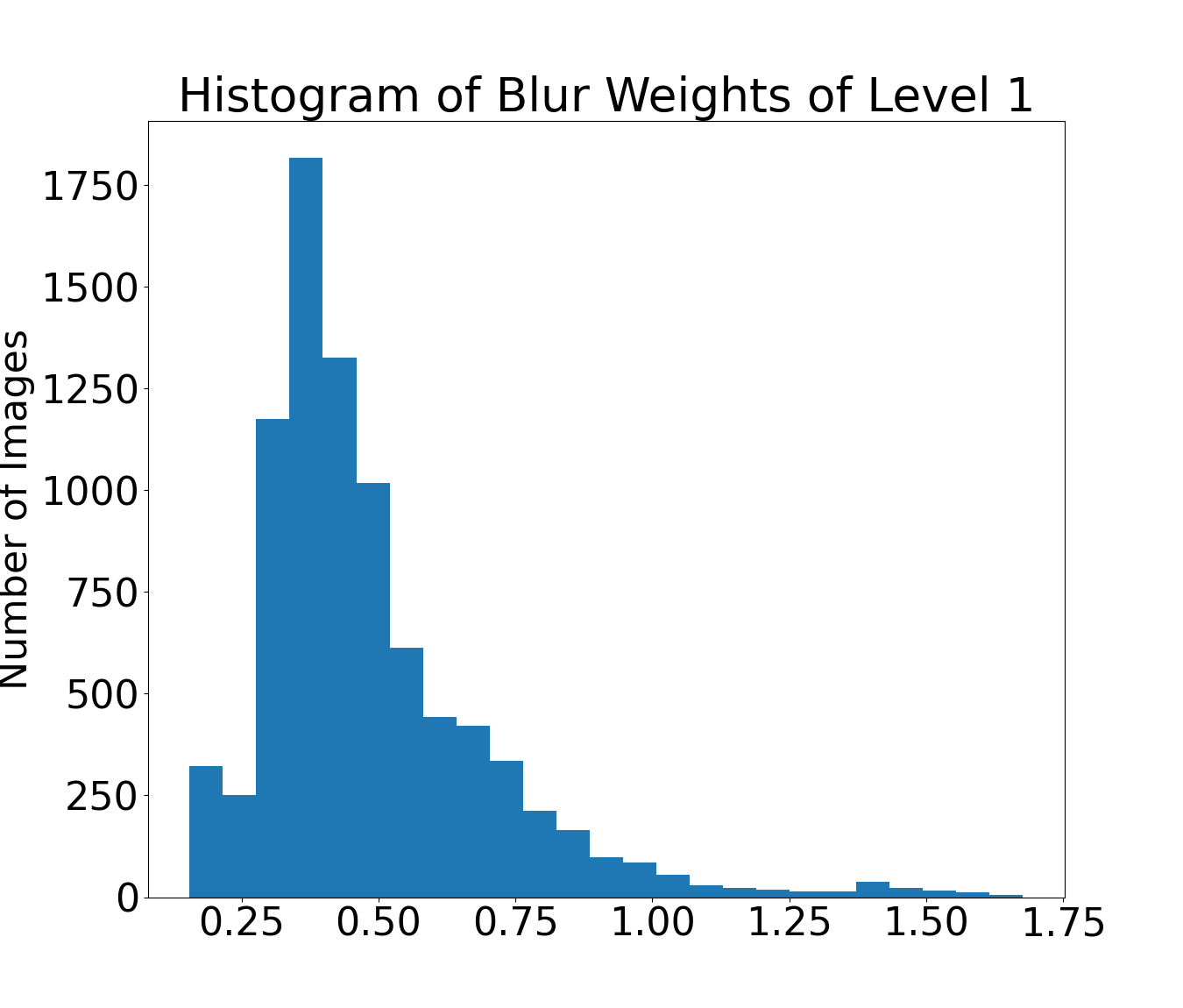}
    \end{subfigure}
    \begin{subfigure}{0.24\linewidth}
        \includegraphics[width=1\linewidth]{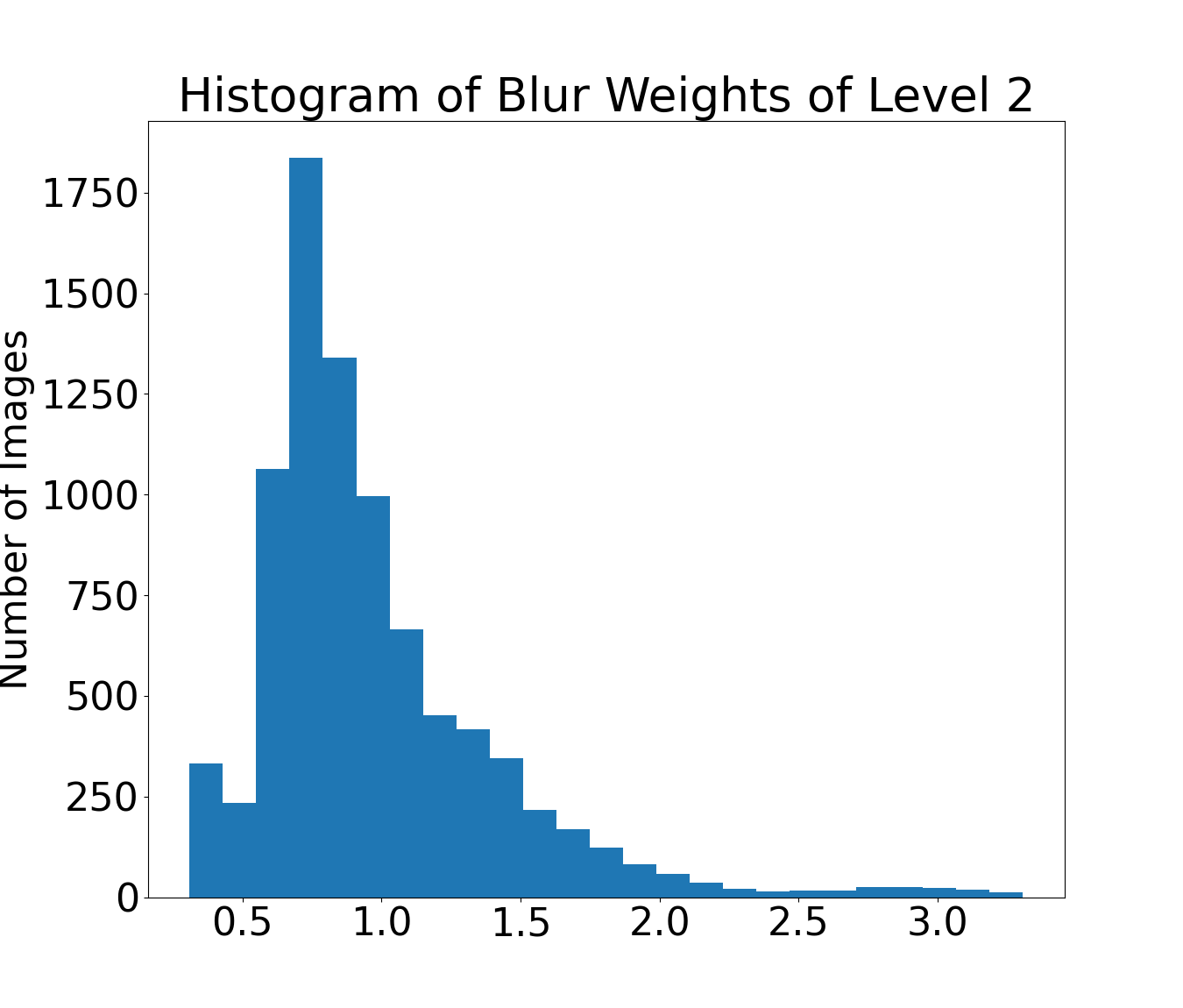}
    \end{subfigure}
    \begin{subfigure}{0.24\linewidth}
        \includegraphics[width=1\linewidth]{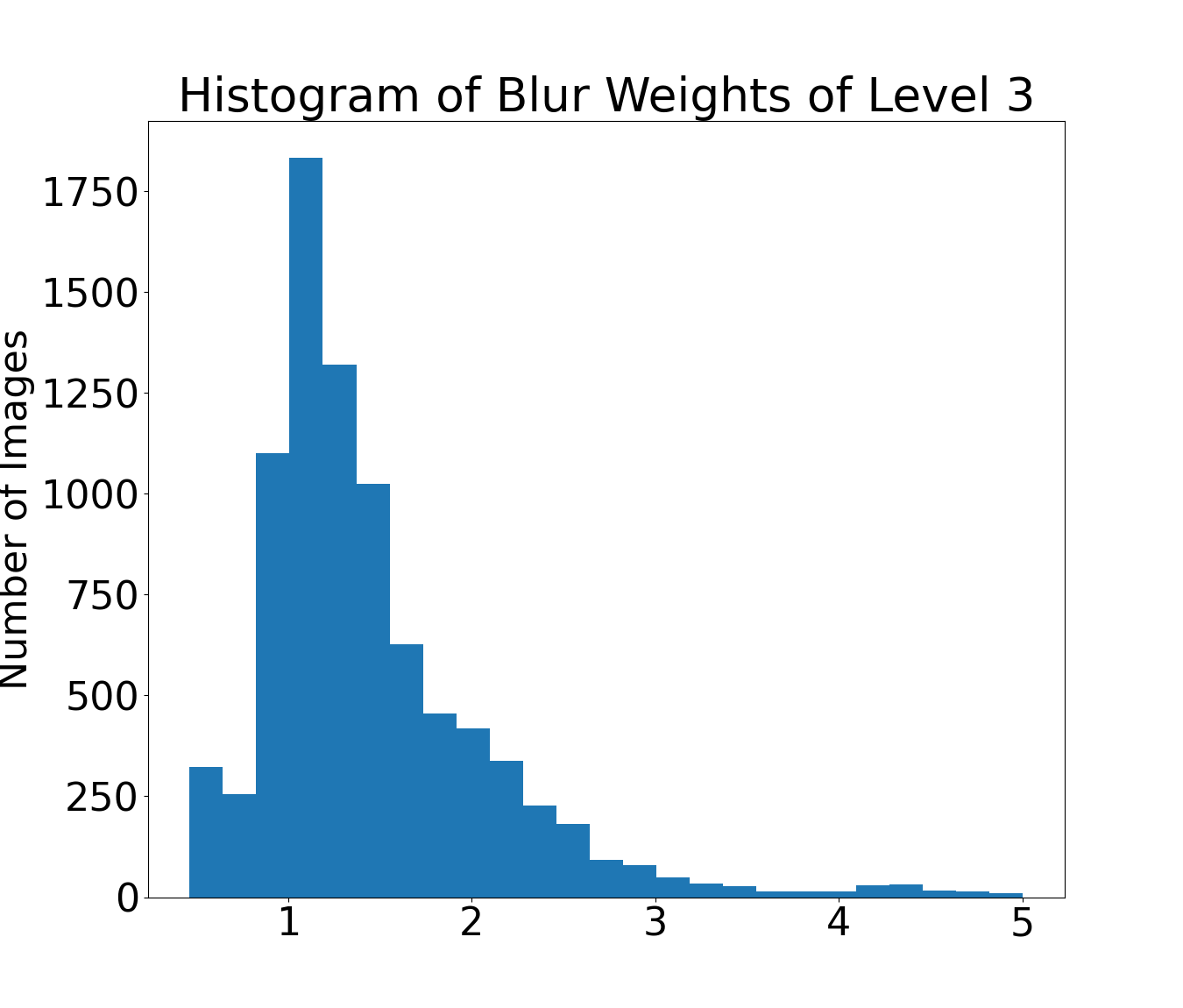}
    \end{subfigure}
    \begin{subfigure}{0.24\linewidth}
        \includegraphics[width=1\linewidth]{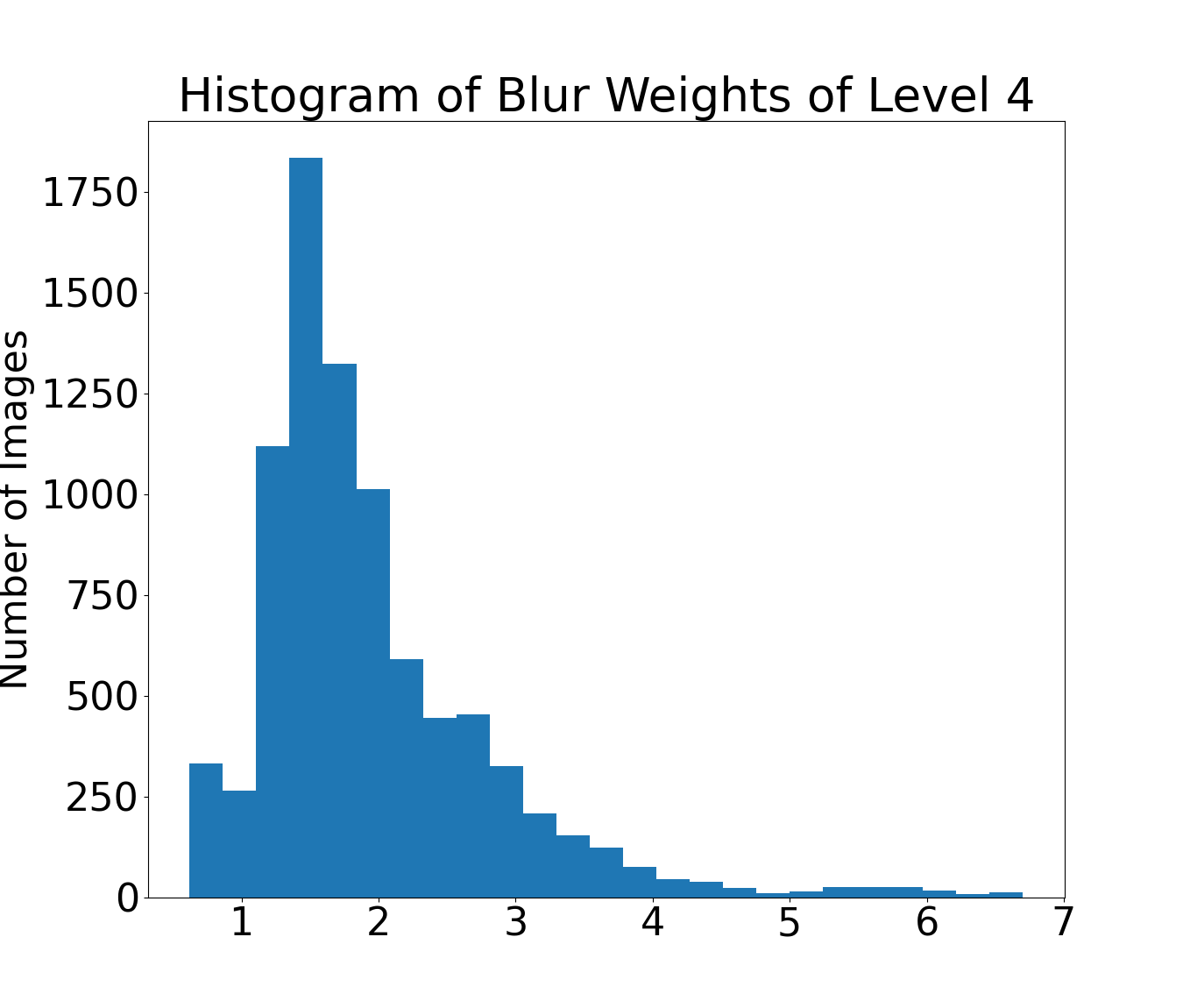}
    \end{subfigure}
    \label{fig:histogram_images}
\end{figure*}

~\cref{fig:teaser} includes additional examples of our rendered images. For a better overview of the dataset,~\cref{fig:histogram_scene,fig:histogram_images} include the histogram of the scene bound, scene dimension, and blur weight for each blur level. Here, scene range is the ratio between the near and far bound of the scene, and \textit{scene dimension} is defined as the geometric mean of the three dimensions of the scene. As a dataset aiming to train a degradation robust Generalizable NeRF, more extensive variety and complexity should result in better generalizability to various use-cases. Additionally, we also show the distribution of blur weights, which is denoted as the $w_l$ of Algorithm~\textcolor{red}{1}. We can notice that our dataset contains a wide variety of scenes with a well-distributed depth range and dimensions. 

\section{Training Details}
In this section, we provide a more comprehensive description of the training and reported results. For training GNeRF models ~\cite{johari2022geonerf, pearl2022nan} with and without our proposed module, we have used an RTX3090 graphic card. During the pretraining stage of the GNeRF models with noise degradation (~\cref{table:num_src_imgs_result,table:composite_result,table:burst_result}), we first apply inverse gamma correction and random white balancing as in ~\cite{pearl2022nan,bhat2021deeprep, xia2020bpn, midhenall2018kpn} to linearize the image, then added noise. The predicted image by the GNeRF renderer is then post-processed by applying the gamma correction and white-balancing for calculating the photometric loss $\mathcal{L}_{\text{photo}}$. With such a setting, the models are trained over 150K and 200K iterations respectively during pretraining stage where the batch size is 512 rays.

For the adversarial training ~\cite{fu2023nerfool} of GNT ~\cite{t2023gnt}, we follow their pretraining configuration which is to train over 250K iterations with batch size 512. We have used RTX3090 when source images are clean during pretraining stage (Row 1-2 of ~\cref{table:gnt_adversarial}). For source images with adversarially degradation during the pretraining stage (Row 3-4 of ~\cref{table:gnt_adversarial}), we have used NVIDIA A100 graphics card. The switch was necessary due to the substantial memory demands arising from learning the adversarial perturbation, which is dependent on the gradient of the entire 3D reconstruction process. Especially when depth estimation is integrated into the computation graph, the memory cost increases significantly. 

\section{Visual Results}

\begin{figure*}[t]
    \caption{Visual results of 3D Differentiable Warping through Depth Estimator $\mathcal{D}$}

    \begin{subfigure}{1.0\linewidth}
        \begin{subfigure}{0.24\linewidth}
            \includegraphics[width=1\linewidth]{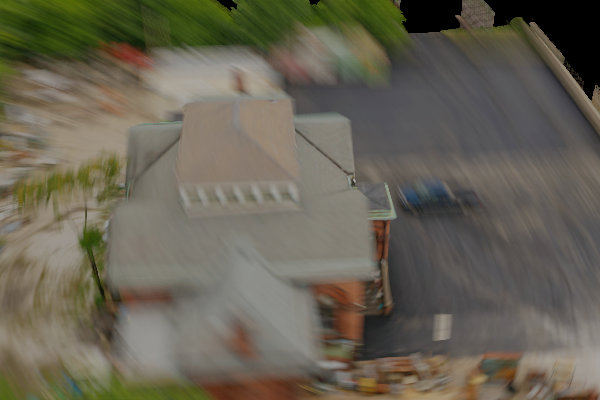}
        \end{subfigure}
        \begin{subfigure}{0.24\linewidth}
            \includegraphics[width=1\linewidth]{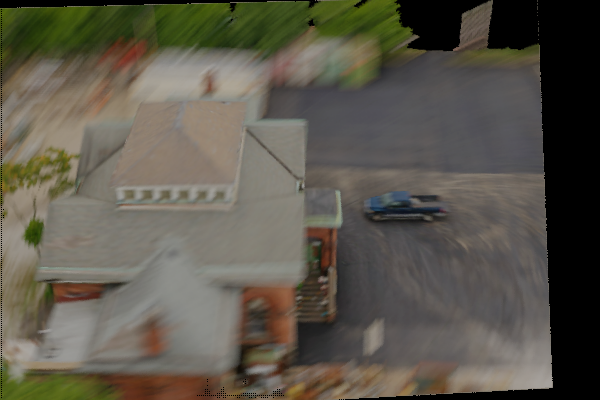}
        \end{subfigure}
        \begin{subfigure}{0.24\linewidth}
            \includegraphics[width=1\linewidth]{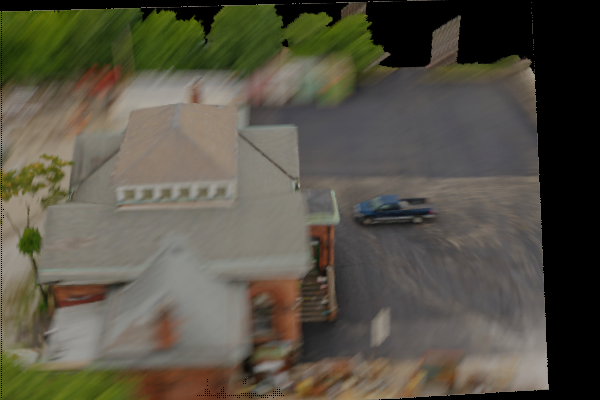}
        \end{subfigure}
        \begin{subfigure}{0.24\linewidth}
            \includegraphics[width=1\linewidth]{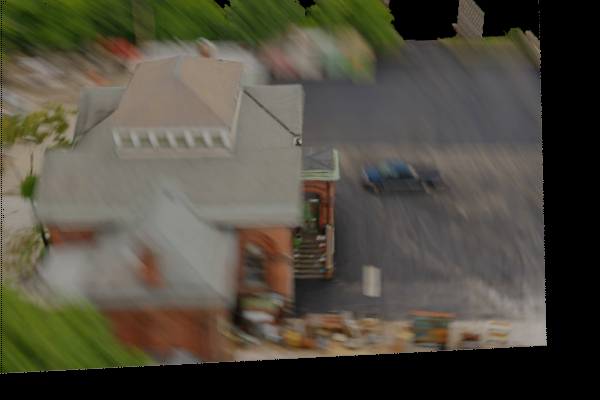}
        \end{subfigure}
    \end{subfigure}

    \begin{subfigure}{1.0\linewidth}
        \begin{subfigure}{0.24\linewidth}
            \includegraphics[width=1\linewidth]{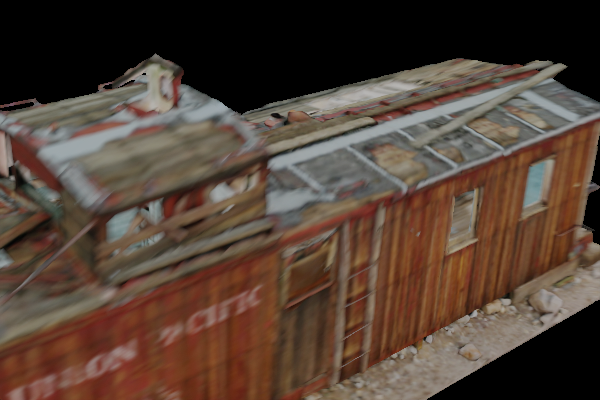}
        \end{subfigure}
        \begin{subfigure}{0.24\linewidth}
            \includegraphics[width=1\linewidth]{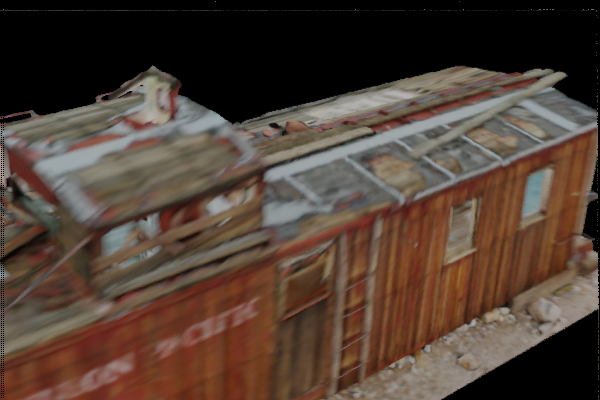}
        \end{subfigure}
        \begin{subfigure}{0.24\linewidth}
            \includegraphics[width=1\linewidth]{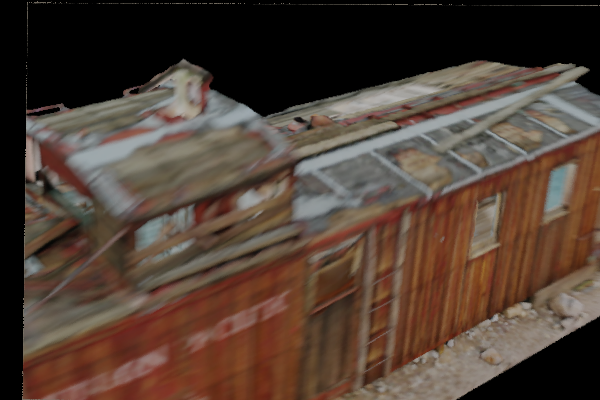}
        \end{subfigure}
        \begin{subfigure}{0.24\linewidth}
            \includegraphics[width=1\linewidth]{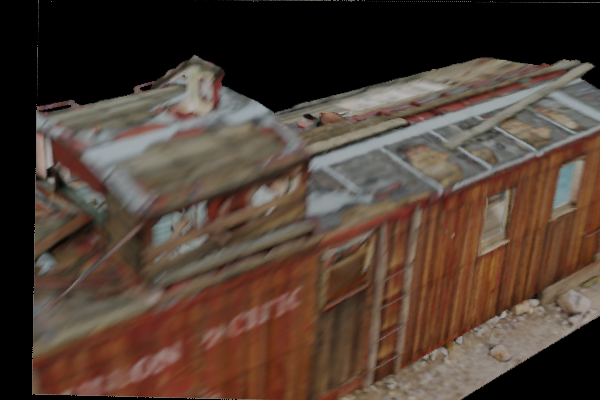}
        \end{subfigure}
    \end{subfigure}

    \begin{subfigure}{1.0\linewidth}
        \begin{subfigure}{0.24\linewidth}
            \includegraphics[width=1\linewidth]{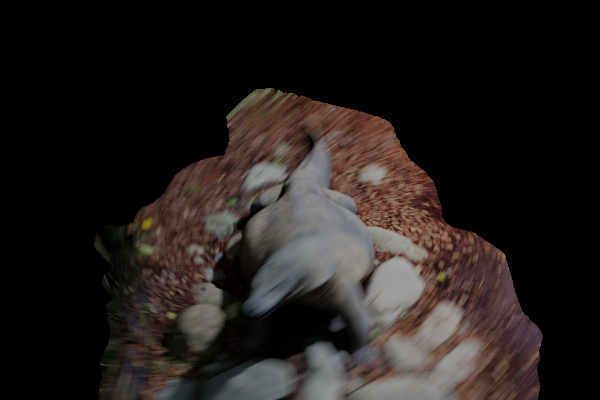}
        \end{subfigure}
        \begin{subfigure}{0.24\linewidth}
            \includegraphics[width=1\linewidth]{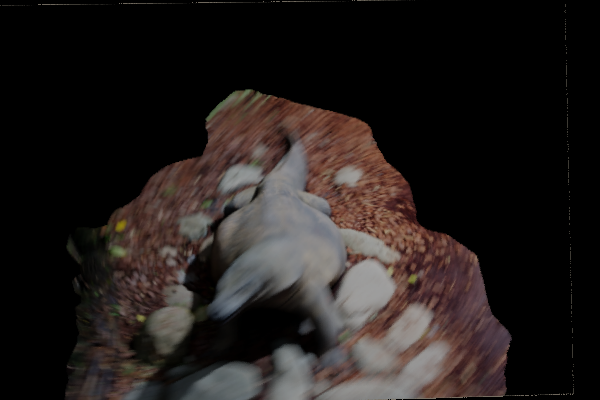}
        \end{subfigure}
        \begin{subfigure}{0.24\linewidth}
            \includegraphics[width=1\linewidth]{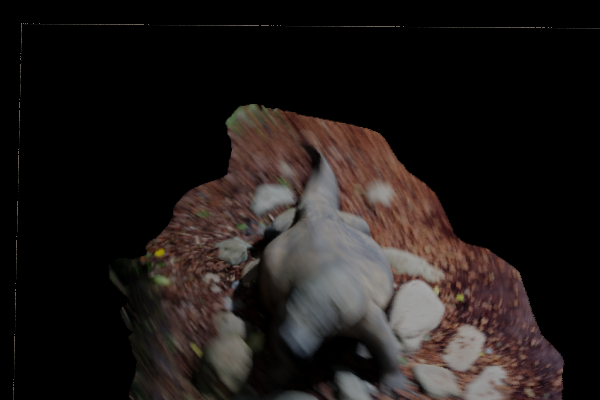}
        \end{subfigure}
        \begin{subfigure}{0.24\linewidth}
            \includegraphics[width=1\linewidth]{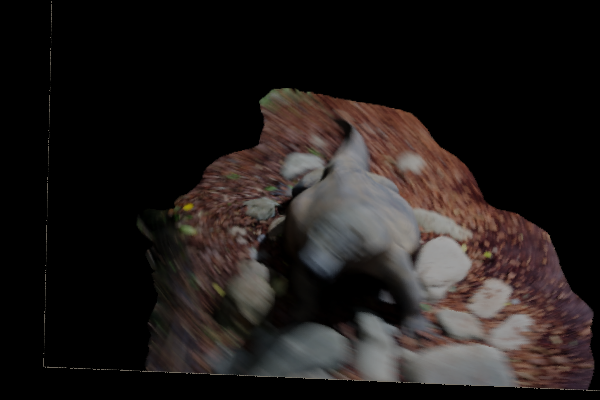}
        \end{subfigure}
    \end{subfigure}

    \begin{subfigure}{1.0\linewidth}
        \begin{subfigure}{0.24\linewidth}\hfil Reference View\end{subfigure}
        \begin{subfigure}{0.72\linewidth}\hfil Nearby Source Views\end{subfigure}
    \end{subfigure}
    \label{fig:supp-depth-module}
\end{figure*}

\begin{figure*}[t]
    \caption{Visual results of 3D-aware Auxiliary Restoration Head $\mathcal{R}$}
    \begin{subfigure}{1.0\linewidth}
        \begin{subfigure}{0.33\linewidth}
            \includegraphics[width=1\linewidth]{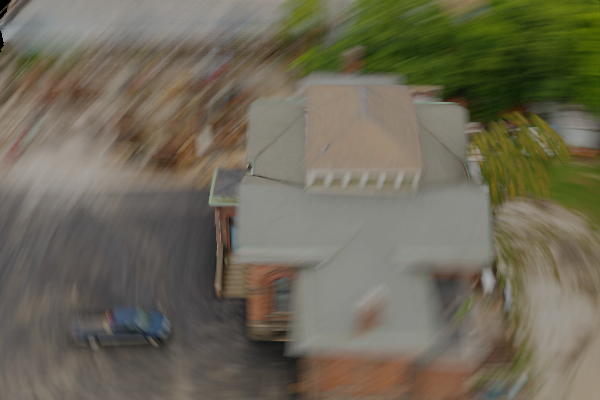}
        \end{subfigure}
        \begin{subfigure}{0.33\linewidth}
            \includegraphics[width=1\linewidth]{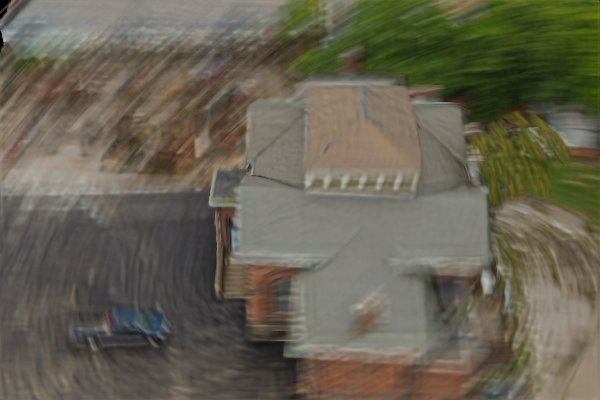}
        \end{subfigure}
        \begin{subfigure}{0.33\linewidth}
            \includegraphics[width=1\linewidth]{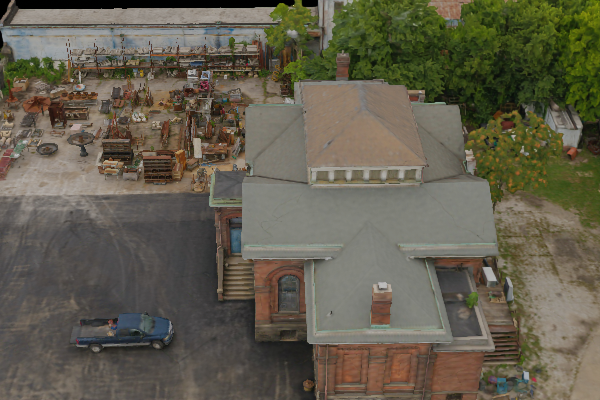}
        \end{subfigure}
    \end{subfigure}

    \begin{subfigure}{1.0\linewidth}
        \begin{subfigure}{0.33\linewidth}
            \includegraphics[width=1\linewidth]{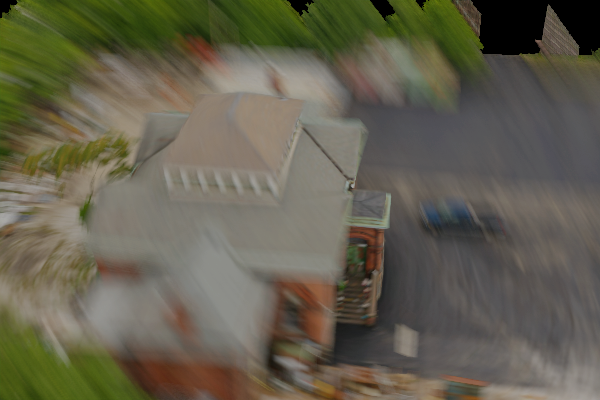}
        \end{subfigure}
        \begin{subfigure}{0.33\linewidth}
            \includegraphics[width=1\linewidth]{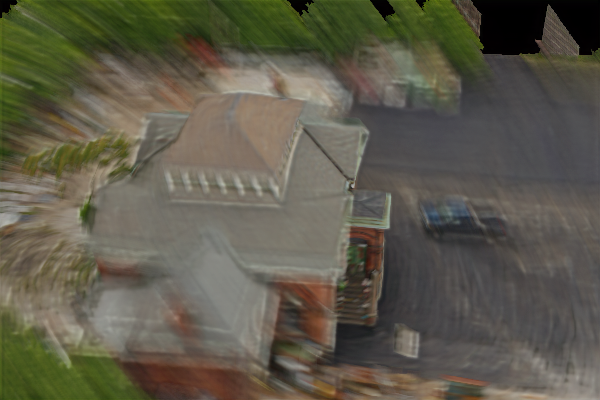}
        \end{subfigure}
        \begin{subfigure}{0.33\linewidth}
            \includegraphics[width=1\linewidth]{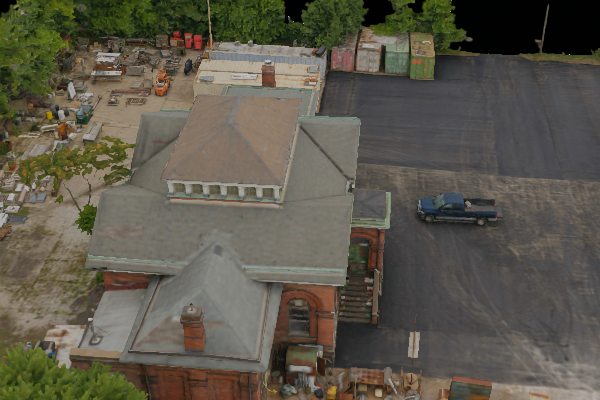}
        \end{subfigure}
    \end{subfigure}

    \begin{subfigure}{1.0\linewidth}
        \begin{subfigure}{0.33\linewidth}
            \includegraphics[width=1\linewidth]{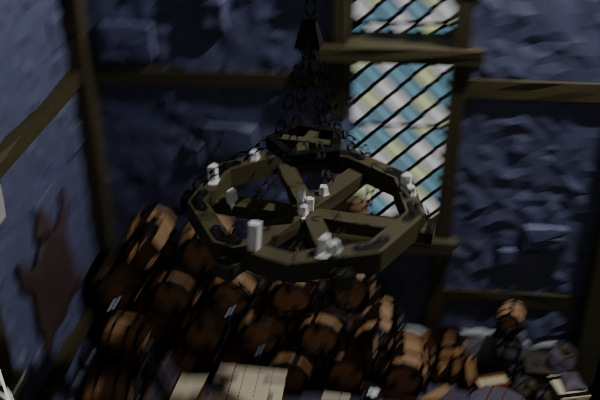}
        \end{subfigure}
        \begin{subfigure}{0.33\linewidth}
            \includegraphics[width=1\linewidth]{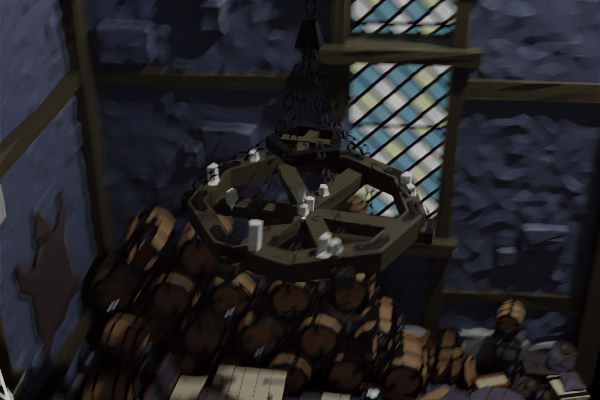}
        \end{subfigure}
        \begin{subfigure}{0.33\linewidth}
            \includegraphics[width=1\linewidth]{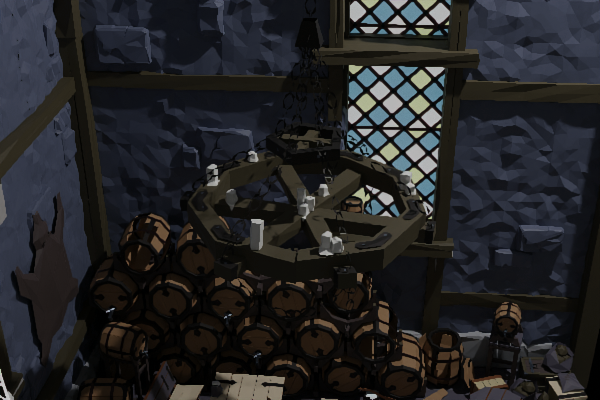}
        \end{subfigure}
    \end{subfigure} 
    
    \begin{subfigure}{1.0\linewidth}
        \begin{subfigure}{0.33\linewidth}
            \includegraphics[width=1\linewidth]{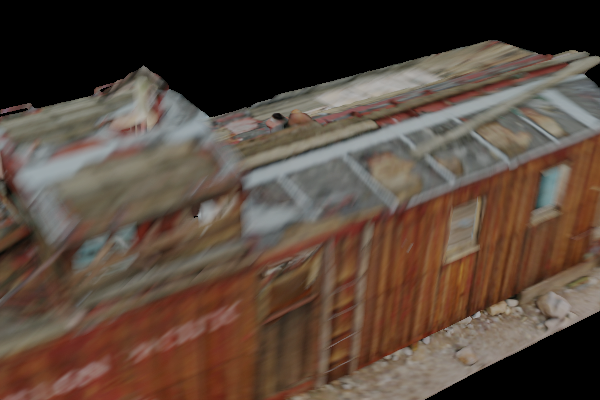}
        \end{subfigure}
        \begin{subfigure}{0.33\linewidth}
            \includegraphics[width=1\linewidth]{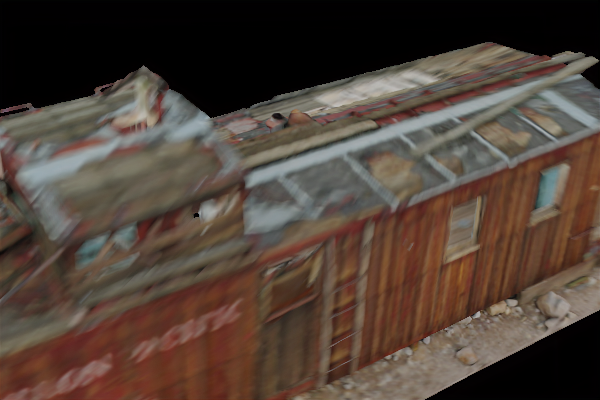}
        \end{subfigure}
        \begin{subfigure}{0.33\linewidth}
            \includegraphics[width=1\linewidth]{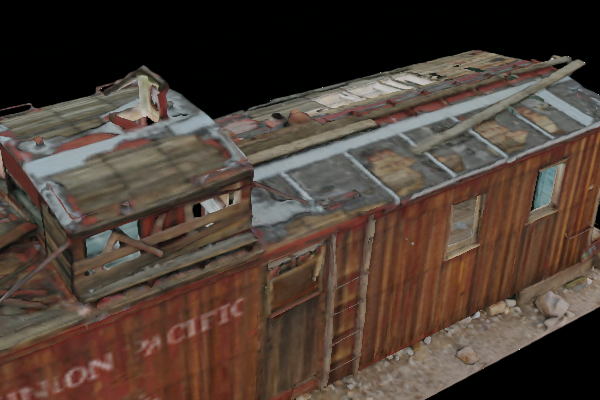}
        \end{subfigure}
    \end{subfigure}
    
    \begin{subfigure}{1.0\linewidth}
        \begin{subfigure}{0.33\linewidth}\hfil Degraded Image\end{subfigure}
        \begin{subfigure}{0.33\linewidth}\hfil Restored Image\end{subfigure}
        \begin{subfigure}{0.33\linewidth}\hfil Ground Truth\end{subfigure}
    \end{subfigure}
    \label{fig:supp-reconst-module}
\end{figure*}

To demonstrate the effectiveness of our proposed 3D-Degradation Aware Feature Extractor, we include visual results of the intermediate outputs. ~\cref{fig:supp-depth-module} visualizes the results of differential warping (Eq.~\textcolor{red}{5})) through the coarse depth predicted by the depth estimator $\mathcal{D}$. This process aims to align degraded source images with each source view, optimizing the extraction of information from the region. By achieving this alignment across multiple views, our method enhances feature awareness among nearby views, thereby improving feature matching and overall image reconstruction.

~\cref{fig:supp-reconst-module} shows the restored image from auxiliary restoration head $\mathcal{R}$, which takes the input of nearby warped image and depths as described in Eq.~\textcolor{red}{6}). By engaging in the auxiliary task of restoring natural signals from degraded source images, our feature extractor adeptly captures the statistical properties inherent in natural images, thereby enhancing the overall 3D reconstruction process.

\section{Quantitative Result}

\cref{table:blur_combined_result,table:adversarial_result,table:composite_result,table:burst_result} include detailed experiment results, including LPIPS and SSIM metrics of 3D reconstruction from degraded source images. Additionally, \cref{table:num_src_imgs_result} is an experiment demonstrating the effect of our proposed module on NAN \cite{pearl2022nan} with 
a different number of source images. Results indicate that the effectiveness of our module increases with the number of source images. Such empirical result suggests that our module is particularly adept at extracting consistent deep image features from multiple degraded sources.

Next, we detail the methodology and underlying rationale presented in Table~\textcolor{red}{5} of the main paper. Utilizing our novel blur dataset, which encompasses varying levels of a blur for identical viewpoints and scenes, we conducted inferences at the same locations under different blur levels to analyze the stability in depth predictions. The results reported are averages across 13 scenes consisting of the 52 test settings. Our dataset is uniquely constructed by averaging latent images, as specified in Algorithm~\textcolor{red}{1}. This approach allows us to emulate realistic blur effects that are depth-dependent, representing a more accurate simulation than the uniform degradation typically achieved with a standard blur kernel.

\section{Limitation}

Our proposed module's current limitation also relates to our future research directions. At present, similar to a predominant number of NeRF-based methods, our approach presupposes the availability of accurate camera poses during both pre-training and inference stages. However, with degraded source images, the reliability of pose estimation methods like COLMAP may be inaccurate. While we have demonstrated our method's effectiveness on real-dataset in Sec.~\textcolor{red}{6.5}, with larger degradation resulting in inaccurate depth estimation and the downstream feature extraction, our method may fail. Recent works, such as those by \cite{lin2021barf, yan2024cfnerf, jiang2023alignerf}, suggest methodologies for adjusting incorrect camera poses during training. Inspired by these advancements, we plan to enhance our module's robustness against misalignment and noise in camera information, aiming to improve its performance under more challenging conditions.

\begin{table*}
\centering
\caption{NAN: Novel view results of different numbers of source views on the \textit{Objavserse Blur} dataset across blur degradation.}
\label{table:num_src_imgs_result}
\begin{tabular}{|c|c|c|c|c|c|c|c|c|}
\hline
Blur Level & Method & \# Source & PSNR $\uparrow$ & $\Delta$ PSNR & SSIM $\uparrow$ & $\Delta$ SSIM & LPIPS $\downarrow$ & $\Delta$ LPIPS \\
\hline
\multirow{6}{*}{1} 
 & \multirow{3}{*}{NAN} 
 &    3 & 23.15 & - & 0.766 & - & 0.268 & -  \\
 & &  5 & 25.18 & - & 0.810 & - & 0.241 & -  \\
 & &  7 & 25.59 & - & 0.823 & - & 0.231 & -  \\
\cline{2-9}
 & \multirow{3}{*}{NAN+Proposed} 
 &    3 & 23.74 & 0.59 & 0.797 & 0.031 & 0.226 & -0.042  \\
 & &  5 & 25.65 & 0.47 & 0.839 & 0.029 & 0.199 & -0.041  \\
 & &  7 & 26.51 & 0.92 & 0.850 & 0.027 & 0.194 & -0.037  \\
\hline
\multirow{6}{*}{2} 
 & \multirow{3}{*}{NAN} 
 &    3 & 21.79 & - & 0.705 & - & 0.328 & -  \\
 & &  5 & 23.71 & - & 0.751 & - & 0.299 & -  \\
 & &  7 & 24.19 & - & 0.766 & - & 0.288 & -  \\
\cline{2-9}
 & \multirow{3}{*}{NAN+Proposed} 
 &    3 & 22.28 & 0.49 & 0.730 & 0.025 & 0.299 & -0.029  \\
 & &  5 & 24.18 & 0.47 & 0.774 & 0.023 & 0.271 & -0.028  \\
 & &  7 & 24.96 & 0.77 & 0.787 & 0.021 & 0.264 & -0.024  \\
\hline
\multirow{6}{*}{3} 
 & \multirow{3}{*}{NAN} 
 &    3 & 21.0 & - & 0.671 & - & 0.363 & -  \\
 & &  5 & 22.82 & - & 0.715 & - & 0.334 & -  \\
 & &  7 & 23.28 & - & 0.731 & - & 0.323 & -  \\
\cline{2-9}
 & \multirow{3}{*}{NAN+Proposed} 
 &    3 & 21.43 & 0.43 & 0.692 & 0.021 & 0.339 & -0.024  \\
 & &  5 & 23.26 & 0.44 & 0.735 & 0.020 & 0.313 & -0.021  \\
 & &  7 & 23.93 & 0.65 & 0.748 & 0.017 & 0.3038 & -0.0192  \\
\hline
\multirow{6}{*}{4} 
 & \multirow{3}{*}{NAN} 
 &    3 & 20.43 & - & 0.649 & - & 0.386 & -  \\
 & &  5 & 22.17 & - & 0.691 & - & 0.357 & -  \\
 & &  7 & 22.63 & - & 0.707 & - & 0.347 & -  \\
\cline{2-9}
 & \multirow{3}{*}{NAN+Proposed} 
 &    3 & 20.84 & 0.41 & 0.667 & 0.018 & 0.365 & -0.021  \\
 & &  5 & 22.58 & 0.41 & 0.708 & 0.017 & 0.339 & -0.018  \\
 & &  7 & 23.16 & 0.53 & 0.721 & 0.014 & 0.330 & -0.017  \\
\hline
\end{tabular}

\end{table*}

\begin{table*} 
\small
\centering
\caption{NAN and GeoNeRF: novel view results on the \textit{Objavserse Blur} dataset across different blur levels.}
\label{table:blur_combined_result}
\begin{tabular}{|l|l|c|c|c|l|c|c|c|}
\hline
Blur & Method & PSNR $\uparrow$ & LPIPS $\downarrow$ & SSIM $\uparrow$ & Method & PSNR $\uparrow$ & LPIPS $\downarrow$ & SSIM $\uparrow$ \\
\hline
\multirow{4}{*}{1}
& GeoNeRF* & 26.28 & 0.279 & 0.779 & NAN* & 23.73 & 0.266 & 0.771 \\
& GeoNeRF & 27.96 & 0.275 & 0.786 & NAN & 25.59 & 0.231 & 0.823 \\
& + Proposed & \textbf{28.47 (+0.51)} & \textbf{0.253 (-0.022)} & \textbf{0.838 (+0.052)} & + Proposed & \textbf{26.51 (+0.92)} & \textbf{0.194 (-0.037)} & \textbf{0.850 (+0.027)} \\
\hline
\multirow{4}{*}{2}
& GeoNeRF* & 24.84 & 0.345 & 0.721 & NAN* & 22.11 & 0.323 & 0.704 \\
& GeoNeRF & 26.17 & 0.338 & 0.727 & NAN & 24.19 & 0.288 & 0.766 \\
& + Proposed & \textbf{26.73 (+0.56)} & \textbf{0.318 (-0.020)} & \textbf{0.776 (+0.049)} & + Proposed & \textbf{24.96 (+0.77)} & \textbf{0.264 (-0.024)} & \textbf{0.787 (+0.021)} \\
\hline
\multirow{4}{*}{3}
& GeoNeRF* & 23.79 & 0.385 & 0.682 & NAN* & 21.14 & 0.353 & 0.671 \\
& GeoNeRF & 25.12 & 0.376 & 0.687 & NAN & 23.28 & 0.323 & 0.731 \\
& + Proposed & \textbf{25.69 (+0.57)} & \textbf{0.357 (-0.019)} & \textbf{0.734 (+0.047)} & + Proposed & \textbf{23.93 (+0.65)} & \textbf{0.304 (-0.019)} & \textbf{0.748 (+0.017)} \\
\hline
\multirow{4}{*}{4}
& GeoNeRF* & 22.91 & 0.415 & 0.645 & NAN* & 20.39 & 0.374 & 0.650 \\
& GeoNeRF & 24.42 & 0.403 & 0.657 & NAN & 22.63 & 0.347 & 0.707 \\
& + Proposed & \textbf{24.977 (+0.56)} & \textbf{0.384 (-0.019)} & \textbf{0.706 (+0.049)} & + Proposed & \textbf{23.16 (+0.53)} & \textbf{0.330 (-0.017)} & \textbf{0.721 (+0.014)} \\
\hline
\end{tabular}
\end{table*}

\begin{table*}
\centering
\caption{Novel view reconstruction results of NAN on noise degradation (\textit{LLFF-N} dataset) across different gain levels.}
\begin{tabular}{|c|c|c|c|c|c|}
\hline
Metric & Method & Gain 4 & Gain 8 & Gain 16 & Gain 20 \\
\hline
\multirow{3}{*}{PSNR $\uparrow$} & NAN & 23.85 & 23.33 & 22.36 & 21.94 \\
                      & + Proposed & 23.90 & 23.58 & 22.89 & 22.57 \\
                      & \textbf{Difference} & \textbf{+0.05} & \textbf{+0.24} & \textbf{+0.53} & \textbf{+0.63} \\
\hline
\multirow{3}{*}{LPIPS $\downarrow$} & NAN & 0.318 & 0.381 & 0.469 & 0.501 \\
                       & + Proposed & 0.299 & 0.350 & 0.424 & 0.453 \\
                       & \textbf{Difference} & \textbf{-0.019} & \textbf{-0.031} & \textbf{-0.045} & \textbf{-0.048} \\
\hline
\multirow{3}{*}{SSIM $\uparrow$} & NAN & 0.754 & 0.706 & 0.623 & 0.586 \\
                      & + Proposed & 0.774 & 0.740 & 0.681 & 0.655 \\
                      & \textbf{Difference} & \textbf{+0.020} & \textbf{+0.034} & \textbf{+0.058} & \textbf{+0.069} \\
\hline
\end{tabular}
\label{table:burst_result}
\end{table*}

\begin{table*}
    \centering
    \caption{Novel view reconstruction results of GNT~\cite{t2023gnt} on different training and test methods across the \textit{LLFF} dataset under adversarial degradation.}
    \label{table:adversarial_result}
    \begin{tabular}{lllccccccccc}
        \toprule
        Training & Test & Proposed & Average & Fern & Flower & Fortress & Horns & Leaves & Orchids & Room & Trex \\
        \midrule
        Clean & Clean & $\times$ & 23.50 & 22.31 & 24.62 & 28.02 & 24.32 & 18.47 & 17.93 & 27.76 & 22.53 \\
               &       & $\checkmark$ & 24.14 & 22.64 & 25.79 & 28.57 & 24.96 & 19.16 & 18.26 & 28.11 & 23.70 \\
               &       & Difference & +0.64 & +0.33 & +1.17 & +0.55 & +0.64 & +0.69 & +0.33 & +0.35 & +1.18 \\
        \hline
        Clean & Adversarial & $\times$ & 17.68 & 18.06 & 18.95 & 20.78 & 18.47 & 15.34 & 14.38 & 19.57 & 15.90 \\
               &             & $\checkmark$ & 18.90 & 18.91 & 20.26 & 21.63 & 19.80 & 16.64 & 15.02 & 20.89 & 17.99 \\
               &             & Difference & +1.22 & +0.85 & +1.31 & +0.85 & +1.33 & +1.30 & +0.64 & +1.32 & +2.09 \\
        \hline       

        Adversarial & Clean & $\times$ & 24.26 & 22.44 & 25.37 & 27.93 & 24.52 & 19.04 & 18.15 & 27.87 & 23.25 \\
                     &       & $\checkmark$ & 24.40 & 22.56 & 25.57 & 27.51 & 24.69 & 19.01 & 18.08 & 28.00 & 23.50 \\
                     &       & Difference & +0.14 & +0.12 & +0.20 & -0.41 & +0.16 & -0.04 & -0.07 & +0.13 & +0.25 \\
        \hline       
        Adversarial & Adversarial & $\times$ & 20.05 & 19.95 & 21.77 & 23.20 & 21.35 & 17.48 & 16.32 & 23.41 & 20.47 \\
                     &            & $\checkmark$ & 20.22 & 20.15 & 22.02 & 23.89 & 21.40 & 17.29 & 16.22 & 23.81 & 20.28 \\
                     &            & Difference & +0.17 & +0.20 & +0.25 & +0.69 & +0.05 & -0.19 & -0.10 & +0.40 & -0.19 \\
        \bottomrule
    \end{tabular}
    \label{table:gnt_adversarial}

\end{table*}

\begin{table*}
\centering
\caption{Novel view reconstruction results of NAN~\cite{pearl2022nan} on the \textit{Objavserse Blur} dataset across different blur levels and burst gain levels.}
\label{table:composite_result}
\begin{tabular}{|c|c|c|c|c|c|c|c|c|}
\hline
Blur Level & Method & Burst Noise Level & PSNR $\uparrow$ & \begin{tabular}[c]{@{}c@{}}$\Delta$ PSNR\end{tabular} & LPIPS $\downarrow$& \begin{tabular}[c]{@{}c@{}}$\Delta$ LPIPS\end{tabular} & SSIM $\uparrow$& \begin{tabular}[c]{@{}c@{}}$\Delta$ SSIM\end{tabular} \\
\hline
\multirow{6}{*}{1} 
 & \multirow{3}{*}{NAN} 
 &  1 & 23.59 & - & 0.274 & - & 0.774 & - \\
 & &  2 & 23.47 & - & 0.293 & - & 0.751 & - \\
 & &  4 & 23.07 & - & 0.336 & - & 0.704 & - \\
\cline{2-9}
 & \multirow{3}{*}{NAN+Proposed} 
 &  1 & 24.28 & +0.69 & 0.244 & -0.029 & 0.802 & +0.028 \\
 & &  2 & 24.21 & +0.74 & 0.257 & -0.036 & 0.792 & +0.041 \\
 & &  4 & 23.97 & +0.90 & 0.293 & -0.043 & 0.762 & +0.059 \\
\hline
\multirow{6}{*}{2} 
 & \multirow{3}{*}{NAN} 
 &  1 & 22.26 & - & 0.328 & - & 0.712 & - \\
 & &  2 & 22.20 & - & 0.342 & - & 0.695 & - \\
 & &  4 & 21.91 & - & 0.376 & - & 0.650 & - \\
\cline{2-9}
 & \multirow{3}{*}{NAN+Proposed} 
 &  1 & 22.70 & +0.44 & 0.308 & -0.019 & 0.733 & +0.021 \\
 & &  2 & 22.65 & +0.45 & 0.316 & -0.026 & 0.725 & +0.031 \\
 & &  4 & 22.39 & +0.48 & 0.343 & -0.033 & 0.698 & +0.049 \\
\hline
\multirow{6}{*}{3} 
 & \multirow{3}{*}{NAN} 
 &  1 & 21.48 & - & 0.359 & - & 0.675 & - \\
 & &  2 & 21.43 & - & 0.371 & - & 0.659 & - \\
 & &  4 & 21.20 & - & 0.400 & - & 0.617 & - \\
\cline{2-9}
 & \multirow{3}{*}{NAN+Proposed} 
 &  1 & 21.80 & +0.32 & 0.347 & -0.012 & 0.692 & +0.017 \\
 & &  2 & 21.73 & +0.30 & 0.352 & -0.019 & 0.685 & +0.026 \\
 & &  4 & 21.41 & +0.21 & 0.373 & -0.027 & 0.662 & +0.045 \\
\hline
\multirow{6}{*}{4} 
 & \multirow{3}{*}{NAN} 
 &  1 & 20.93 & - & 0.379 & - & 0.651 & - \\
 & &   2 & 20.88 & - & 0.391 & - & 0.635 & - \\
 & &   4 & 20.66 & - & 0.417 & - & 0.596 & - \\
\cline{2-9}
 & \multirow{3}{*}{NAN+Proposed} 
 &  1 & 21.17 & +0.24 & 0.370 & -0.009 & 0.666 & +0.015 \\
 & &   2 & 21.09 & +0.21 & 0.374 & -0.017 & 0.660 & +0.025 \\
 & &   4 & 20.74 & +0.08 & 0.392 & -0.025 & 0.638 & +0.042 \\
\hline
\end{tabular}
\end{table*}